%% file: main.tex
\PassOptionsToPackage{table}{xcolor}
\documentclass[10pt,twocolumn,letterpaper]{article}

\usepackage[pagenumbers]{cvpr} % To force page numbers, e.g. for an arXiv version

\input{preamble}

\definecolor{cvprblue}{rgb}{0.21,0.49,0.74}
\usepackage[pagebackref,breaklinks,colorlinks,citecolor=cvprblue]{hyperref}

\def\paperID{16571} % *** Enter the Paper ID here
\def\confName{CVPR}
\def\confYear{2024}

\title{DAS: A Deformable Attention to Capture Salient Information in CNNs}

\author{\parbox{\textwidth}{\centering
    Farzad Salajegheh $^{1}$ \; \hspace{-10pt}
    \qquad Nader Asadi $^{2}$\,\thanks{Equal contribution, name order randomized.\\Correspondence to \texttt{farzad.salajegheh@concordia.ca, mudur@cs.concordia.ca}} \hspace{-10pt}
    \qquad Soroush Saryazdi $^{3 \; *}$ \hspace{-10pt}
    \qquad Sudhir Mudur $^{1}$}\vspace{10pt}\\
$^1$ Concordia University ~ $^2$ Mila - Quebec AI Institute  ~ $^3$ Matic\\
}

\begin{document}
\maketitle

\begin{abstract}

    Convolutional Neural Networks (CNNs) excel in local spatial pattern recognition. For many vision tasks, such as object recognition and segmentation, salient information is also present outside CNN’s kernel boundaries. However, CNNs struggle in capturing such relevant information due to their confined receptive fields. Self-attention can improve a model’s access to global information but increases computational overhead. We present a fast and simple fully convolutional method called DAS that helps focus attention on relevant information. It uses deformable convolutions for the location of pertinent image regions and separable convolutions for efficiency. DAS plugs into existing CNNs and propagates relevant information using a gating mechanism. Compared to the O(n$^{\text{2}}$) computational complexity of transformer-style attention, DAS is O(n). Our claim is that DAS's ability to pay increased attention to relevant features results in performance improvements when added to popular CNNs for Image Classification and Object Detection. For example, DAS yields an improvement on Stanford Dogs (4.47\%), ImageNet (1.91\%), and COCO AP (3.3\%) with base ResNet50 backbone. This outperforms other CNN attention mechanisms while using similar or less FLOPs. Our code will be publicly available.
    
    % Convolutional Neural Networks (CNNs) excel in local spatial pattern recognition. Yet, they struggle to capture global relations due to their confined receptive fields. Self-attention can improve a model's access to global information but increases computational overhead. We present a fast and simple fully convolutional attention method called DAS. DAS integrates into existing CNNs and propagates relevant information using a gating mechanism. Compared to $O(n^2)$ transformer-style attention, DAS is $O(n)$ and uses separable convolutions for efficiency. We show that adding DAS to popular CNNs for Image Classification and Object Detection improves performance. For example, DAS yields an improvement on Stanford Dogs (4.47\%), ImageNet (1.91\%), and COCO AP (3.3\%) with base ResNet-50 backbone. This outperforms other CNN attention mechanisms while using similar or less FLOPs. Our code will be publicly available.
\end{abstract}

\input{sections/introduction}

\input{sections/method}
\input{sections/experiments}

\input{sections/visualizations}
% \input{sections/ablation}

%\vspace{30pt}

\section{Conclusion, Limitations and Extensions}\label{sec:conclusion}

In this paper, we presented the DAS attention gate, a new self-attention mechanism for CNNs. DAS does not make use of transformers. Compared to earlier methods for attention within CNNs, DAS provides dense attention and looks holistically at the feature context. DAS is very simple – it combines depthwise separable convolutions (for efficient representation of the global context) and deformable convolutions (for increasing focus on pertinent image regions). Implementation results indeed show that DAS, though simple, enables focused attention to task-relevant features in an image. In our view, its simplicity is its power, as (i) it can be introduced between any two layers of a CNN designed for any visual task, (ii)  does not require any change to the rest of the network, (iii) provides dense attention, (iv) provides attention in a holistic fashion, not separating channel or spatial attention, (v) has just a single additional hyper-parameter, that is easy to tune, (vi) adds only a small amount of computation overhead, (vii) is $O(n)$ as opposed to Transformer-style self-attention's $O(n^2)$, and (viii) yields, as of today, the best results as compared to all other earlier proposed CNN attention methods.
% Attention powers some of the most exciting large AI models even for visual tasks.
% The transformer paradigm used is fundamentally a quadratic operation and limits use in resource constrained situations. We propose the CNN - DAS combination as a very good alternative in these situations.

One limitation is that the computation overhead can increase significantly when the network has large depth features. Hence the value of $\alpha$  has to be chosen carefully. Too small a value will result in loss of contextual information and a large value will increase the amount of computation.

While we have demonstrated DAS’s performance for Image Classification and Object Detection, in the future we want to use it for dense vision tasks such as semantic segmentation and stereo matching where DAS’s dense attention capability could offer significant advantages.
 % In the slightly longer term, we would like to take advantage of DAS's "no change in backbone"  feature, and investigate its addition to small and large pretrained CNNs in the context of transfer learning or few shot learning, and devise an appropriate training strategy for effective performance. 

{
    \small
    \bibliographystyle{ieeenat_fullname}
    \bibliography{main}
}


\end{document}

%% file: preamble.tex
\usepackage{microtype}
\usepackage{graphicx}
\usepackage{booktabs}
\usepackage{multirow}
\usepackage{adjustbox}

\usepackage{amsmath}
\usepackage{amssymb}
\usepackage{amsthm}
\usepackage{mathtools}

\usepackage{pgfplots}
\usepackage{colortbl}

\usepackage[table]{xcolor}

\colorlet{shadecolor}{gray!20}

%% file: sections/introduction.tex
\section{Introduction}\label{sec:intro}
Convolutional Neural Networks (CNNs) are architecturally designed to exploit local spatial hierarchies through the application of convolutional filters realized using kernels. While this makes them efficient and effective for tasks that involve local spatial patterns, their intrinsic design restricts their receptive field, and can impede the full integration of relevant information not within the kernel boundaries. Vision Transformers (ViT) \cite{dosovitskiy2020image} support capturing of global dependencies and contextual understanding in images, and are showing improved performance in many computer vision tasks. ViTs decompose images into sequences of flattened patches and subsequently map them to embedding vector sequences for the Transformer encoder. This patch-based approach is adopted due to the attention mechanism's inherent $O(n^2)$ computational complexity with respect to the number of input vectors. By converting the image into coarser patches, ViTs effectively reduce the number of input patches, \textit{i.e.} $n$. However, affording dense attention at granularities, say pixel-wise, remains computationally challenging. Further, ViTs tend to require larger model sizes, higher memory requirements, and extensive pretraining compared to CNNs, and their computational demands limit their practicality in real-time embedded applications. While efforts continue to contain the quadratic complexity of transformers to enable dense attention using convolutions on long sequences \cite{poli2023hyena}, there is considerable research \cite{guo2022attention} in incorporating self-attention mechanisms directly into CNNs with the goal of providing dense salient feature attention. This work is primarily motivated by the latter.
% They have significantly enhanced the capabilities of CNNs by providing mechanisms for capturing contextual information.

% \begin{figure}[t]
%     \centering
%     \includegraphics[width=8.5cm]{assets/DAS_1.png}
%     \caption{DAS attention combines depthwise separable convolution with deformable sampling to find long range dependencies and calculates pixel-wise attention weights}
%     \label{fig:simpleoverview}
% \end{figure}

\begin{figure*}[t]
    \centering
    \resizebox{0.99\textwidth}{!}{\includegraphics{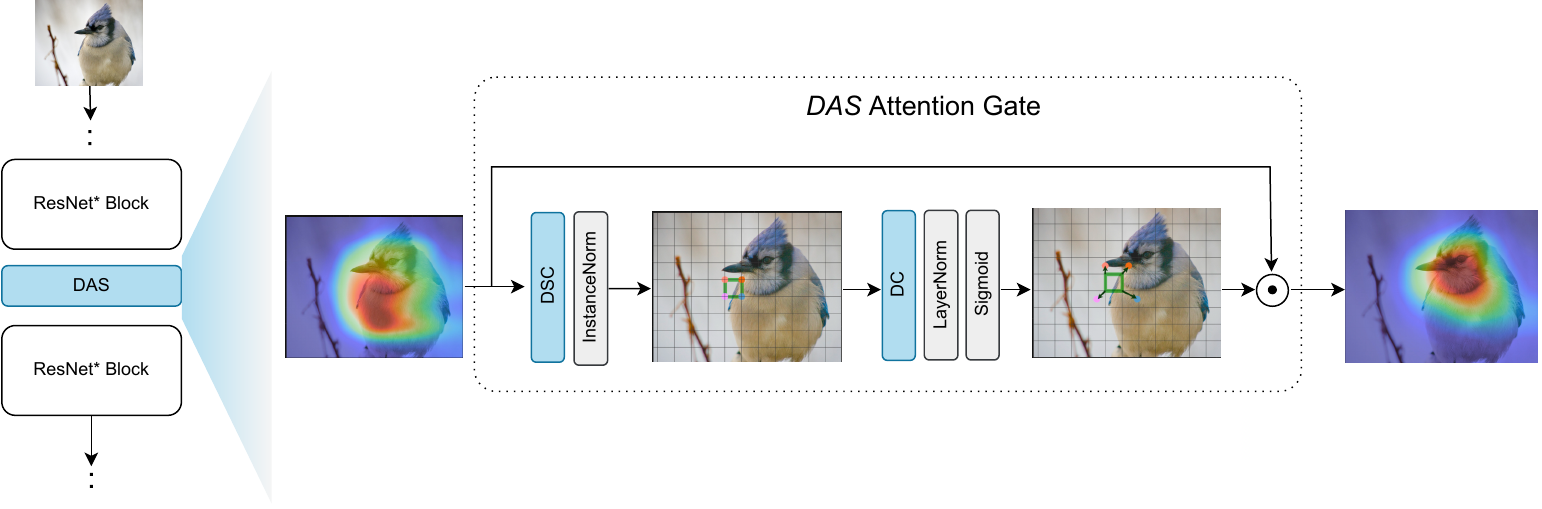}}
    \caption{DAS attention integrates depthwise separable convolution (DSC) and deformable convolution (DC) to focus and increase attention to salient regions, and computes dense attention (pixel-wise) weights. In this figure, the leftmost heatmap shows the ResNet-50 saliency map without attention (shown here for illustration only) and the rightmost shows the same layer, but after DAS gating.}
    \label{fig:simpleoverview}
\end{figure*}

% Attention mechanisms in CNNs can be broadly categorized into channel attention, spatial attention, and mixed-domain attention. Most of the methods present strategies to contain the computations, such as aggregation, subsampling, pooling, etc. For example, most papers that follow the work on stacking attention modules \cite{wang2017residual} resort to using max pooling operations before calculating attention weights in the attention-aware feature map. Another popular strategy is to compute one weight per channel \cite{SENet, ECA-Net} . This may result in ignoring essential spatial contextual information. Some methods have been proposed to extend the above by blending channel and spatial attention\cite{CBAM, BAM}, yielding more robust attention modules. Another extension, \cite{TA} uses the global pooling of two rotations of input along with the global pooling of the original tensor and combines information from three views of the feature. However, they all still grapple with modeling long-range dependencies effectively and treat channel and spatial attention as independent processes. They do not holistically look at the information in a feature, and this could lead to potential information loss.

Attention mechanisms in CNNs can be broadly categorized into \textit{channel attention}, \textit{spatial attention}, and \textit{mixed-domain attention}. These methods- present strategies to contain attention specific computations, using techniques such as aggregation, subsampling, pooling, etc. which in turn makes it difficult to provide dense attention. For example, most papers that follow the work on stacking attention modules \cite{wang2017residual} resort to using average pooling operations before calculating attention weights in the attention-aware feature map. A popular strategy is to compute one weight per channel \cite{SENet, ECA-Net}. This may result in ignoring essential spatial contextual information. Some methods have been proposed to extend the above by blending channel and spatial attention\cite{CBAM, BAM}, yielding more robust attention modules. Another extension, \cite{TA} uses the global pooling of two rotations of input along with the global pooling of the original tensor and combines information from three views of the feature. However, they all still grapple with providing attention to salient features effectively. They treat channel and spatial attention as independent processes, thereby they do not holistically look at the information in a feature, and this could lead to potential information loss.

One promising avenue for increasing attention to pertinent regions of the image is the use of deformable grids instead of the regular grids that are used in standard convolutional filters. Deformable ConvNets v2 \cite{zhu2019deformable} has shown an improved ability to focus on pertinent image regions. These types of methods \cite{zhu2020deformable, xia2022vision} have been used to provide deformable attention in ViTs for fine-scale tasks of semantic segmentation and image classification by finding better keys and queries in ViTs. However, our primary interest is in providing an attention mechanism directly in CNNs with minimal changes to the original network or its training. Accordingly, the focus of the rest of this paper is on convolutional attention methods. 

Our method is inspired partly by the success of deformable convolutions \cite{zhu2019deformable}, and partly by the dominance of Raft architecture design on a variety of vision tasks such as optical flow \cite{raft} and stereo vision \cite{raftstereo} which propagate the image/feature map recursively using a gated recurrent unit (GRU).  Our main contribution is an efficient gated-attention mechanism, DAS, which focuses and increases attention to salient image regions. It can very easily be integrated into any existing CNN to enhance the CNN's performance with minimal increase in FLOPs, and importantly, with no change in the backbone. Our attention gate combines the context provided by layer features with the deformable convolution's ability to focus on pertinent image regions to elegantly increase attention to salient features (\cref{fig:simpleoverview}). DAS adds just a single hyperparameter which is also easy to tune. We demonstrate the incorporation of our gate into standard CNNs like ResNet \cite{he2016deep} and MobileNetV2 \cite{sandler2018mobilenetv2} and, through extensive experimental results, show performance gains in various tasks. In support of our claim that CNNs with the addition of our attention gate do indeed focus and increase attention on task-relevant features, we show gradCAM~\cite{selvaraju2017grad} heatmap visuals that highlight important pixels. We also define and compute a simple metric called \textit{salient feature detection} (\textit{sfd}) score for quantitatively comparing the effectiveness of our attention gate.

\section{Related Work}\label{sec:background}
CNN attention mechanisms have been developed to eliminate redundant information flowing through the neural network, while simultaneously addressing the problem of computation load. The goal is to increase attention to salient features and pay reduced/no attention to irrelevant features.

\textbf{Channel Attention.}
Squeeze-and-Excitation Networks (SENet) \cite{SENet} introduced an efficient channel-wise attention mechanism using global pooling and fully connected layers. SENet computes a single attention weight for each channel, resulting in significant performance improvements compared to the base architecture. Meanwhile, the Global Second-order Pooling Networks (GSoP-Net) \cite{GSoP-Net} method employs second-order pooling to compute attention weight vectors. Efficient channel attention (ECA-Net) \cite{ECA-Net} computes attention weights for each channel through global average pooling and a 1D convolution.  Spatial contextual information is largely ignored in the above channel-wise attention methods.

\textbf{Spatial Attention.}
GE-Net \cite{GE-Net} spatially encodes information through depthwise convolutions and then integrates both the input and encoded information into the subsequent layer. The Double Attention Networks (A2-Nets) \cite{A2-Nets} method introduces novel relation functions for Non-Local (NL) blocks, utilizing two consecutive attention blocks in succession.
The Global-Context Networks (GC-Net) \cite{GC-Net} method integrates NL-blocks and SE blocks using intricate permutation-based operations to capture long-range dependencies. CC-Net \cite{CC-Net} combines contextual information from pixels along intersecting trajectories. SA-NET \cite{SA-NET} utilizes channel splitting to process sub-features in parallel. In all the above spatial-attention methods, while the goal is more towards capturing long-range dependencies, computation overhead can be high, as can be seen in our experimental results as well. 

\textbf{Channel-Spatial Attention.}
The Convolutional Block Attention Module (CBAM) \cite{CBAM} and Bottleneck Attention Module (BAM) \cite{BAM} separate channel and spatial attentions and combine them in the last step, to yield better performance than SENet. CBAM's attention blocks incorporate multi-layer perceptrons (MLP) and convolutional layers, employing a fusion of global average and max pooling. A pooling technique called strip pooling is introduced in SP-Net \cite{SP-Net}, utilizing a long and narrow kernel to effectively capture extensive contextual details for tasks involving pixel-wise prediction.  GALA \cite{GALA} also finds the local and global information separately with two 2D tensors and integrates them to get channel-spatial attentions.   Triplet Attention \cite{TA} captures cross-dimensional interactions by permuting input tensors and pooling, leading to performance enhancements. DRA-Net \cite{DRA-Net} also employs two separate FC layers to capture channel and spatial relationships. OFDet \cite{jin2023hybrid} uses all three, channel, spatial, and channel-spatial attentions simultaneously. In all the above, these separately processed attentions will need to be judiciously combined to provide a more holistic representation of the dependency on the feature. Since averaging and/or pooling are used, providing dense attention is also difficult. Again, the computation overhead is high.

A survey on attention mechanisms in CNNs \cite{guo2022attention} puts them into six categories, channel attention, spatial attention, temporal attention, branch attention, channel \& spatial attention, and spatial \& temporal attention. Our proposed attention module does not separate attentions as above, instead, it looks at the whole feature at once and returns pixel-wise attention weights in a very simple approach.

In summary, existing approaches have not completely addressed capturing of channel, spatial and relevant global dependencies in a holistic manner, which is crucial for understanding contextual information. Dense attention and/or computation overheads can also be a problem in most cases. In contrast, our proposed attention gate combines the strengths of depthwise separable convolution and deformable convolution to holistically provide pixel-wise attention. It enables our model to focus and increase attention to relevant information effectively while maintaining the architectural simplicity of CNNs. 

% \nader{For Methodology: Our attention are used after skip connections of each main block in both ResNet and MobileNetV2 models. 4 for ResNet, 7 for MobileNet. Please change the name of groupnorm to InstanceNorm and layernorm. Separate 6 channels into 6 groups (equivalent with InstanceNorm)
% Put all 6 channels into a single group (equivalent with LayerNorm)}

% While LMA \cite{LMA} suggests average pooling along the x and y axes, which are then combined to derive attention weights.

%% file: sections/method.tex
\section{Methodology}\label{sec:method}

In this section, we present our DAS attention mechanism, designed to enhance the capabilities of CNNs in a computationally efficient way to provide focused attention to relevant information. We illustrate the use of our DAS attention gate by employing it after skip connections of each main block in ResNet~\cite{he2016deep} and MobileNetV2~\cite{sandler2018mobilenetv2} models. The key steps and components of our method are described below.

% \subsection{Preliminaries}
% Here, we briefly revisit the \textit{deformable convolution} operation in neural networks. Similar to regular convolution operation, deformable convolution operates across the channel dimension of a three-dimensional feature map. The 2D deformable convolution consists of, first, sampling over the input feature map $\mathbf{z}$, followed by a weighted summation of sample features by kernel parameters $\mathbf{w}$.

% In deformable convolution, the regular convolution grid is augmented with an offset parameter $\Delta {\mathbf{p}_n}$. So for each location $\mathbf{p}$ of the output feature map $\mathbf{z}$, the convolution operation becomes:
% \begin{equation}
%     \mathbf{z}^{\ell}(p) = \sum_{\mathbf{p}_n \in \mathcal{P}} \mathbf{w}(\mathbf{p}_n) . \mathbf{z}^{\ell - 1}(\mathbf{p} + \mathbf{p}_n + \Delta \mathbf{p}_n)   
% \end{equation}
% where $$ TODO complete this sentence.

\subsection{Bottleneck Layer}
We use a depthwise separable convolution operation that acts as a bottleneck layer. This operation reduces the number of channels in the feature maps, transforming them from $c$ channels to $\alpha \times c$ channels, where $0 < \alpha < 1$ . This size reduction parameter $\alpha$ is selected to balance computational efficiency with accuracy. The optimal value for $\alpha$ was determined empirically through experiments presented in our ablation study (Fig. \ref{fig:alphas}). It also shows that the only hyperparameter ($\alpha$) that is added by our model is not very sensitive for $\alpha>0.1$

After the bottleneck layer, we apply a normalization layer, specifically Instance Normalization, followed by a GELU non-linear activation. These operations enhance the representational power of the features and contribute to the attention mechanism's effectiveness. The choice of Instance and Layer Normalization are supported by experimental results in Table \ref{tab:normalization_layers}. Eq. \ref{eq1} shows the compression process where \textbf{X} is the input feature and $\textbf{W}_1$ represents the depthwise separable convolution.
\begin{equation} \label{eq1}
\mathbf{X}_c = \text{GELU}(\text{InstanceNorm}(\mathbf{X} \mathbf{W}_1))
\end{equation}

In Table \ref{tab:normalization_layers}, we show the importance of using InstanceNorm as the normalization technique before deformable convolution operation. Intuitively, the instance normalization process allows to remove instance-specific contrast information from the image which improves the robustness of deformable convolution attention during training.

\begin{table*}[t!]
  \centering
  \begin{tabular}{ll}
  \begin{tabular}{l cccc}
    \toprule
    \multirow{2}{*}{Method} &  \multicolumn{4}{c}{ImageNet1K} \\
     & Parameters (M) & FLOPs (G) & Top-1 (\%) & Top-5 (\%)\\
    \midrule

    ResNet-18~\cite{he2016deep} & \textbf{11.69}  & \textbf{1.82} & 69.76 & 89.08\\
    + SENet~\cite{SENet} & 11.78 & \textbf{1.82} & 70.59 & 89.78 \\
    + BAM~\cite{BAM} & 11.71 & 1.83 & 71.12 & 89.99 \\
    + CBAM~\cite{CBAM} & 11.78 & \textbf{1.82} & 70.73 & 89.91 \\
    + Triplet Attention~\cite{TA} & \textbf{11.69} & 1.83 & 71.09 & 89.99 \\
    + EMCA~\cite{EMCA} & \textbf{11.19}& \textbf{1.70} & 71.00 & 90.00 \\
    \rowcolor{shadecolor}+ DAS~(ours) & 11.82 & 1.86 & \textbf{72.03} & \textbf{90.70} \\\rowcolor{white}

    \midrule

    ResNet-50~\cite{he2016deep} & \textbf{25.56} & \textbf{4.12} & 76.13 & 92.86 \\
    + SENet~\cite{SENet} & 28.07 & 4.13 & 76.71 & 93.38 \\
    + BAM~\cite{BAM} & 25.92 & 4.21 & 75.98 & 92.82 \\
    + CBAM~\cite{CBAM} & 28.09 & 4.13 & 77.34 & 93.69 \\
    + GSoP-Net~\cite{GSoP-Net} & 28.29 & 6.41 & 77.68 & 93.98 \\
    + A$^\text{2}$-Nets~\cite{A2-Nets} & 33.00 & 6.50 & 77.00 & 93.50 \\
    + GCNet~\cite{GC-Net} & 28.10 & 4.13 & 77.70 & 93.66 \\
    + GALA~\cite{GALA} & 29.40 & - & 77.27 & 93.65 \\
    + ABN~\cite{fukui2019attention} & 43.59 & 7.66 & 76.90 & - \\
    + SRM~\cite{lee2019srm} & 25.62 & \textbf{4.12} & 77.13 & 93.51 \\
    + Triplet Attention~\cite{TA} & \textbf{25.56} & 4.17 & 77.48 & 93.68\\
    % + CAT~\cite{} & 26.51& 3.95 & 77.99 & 94.14 \\
    + EMCA~\cite{EMCA} & \textbf{25.04}& \textbf{3.83} & 77.33 & 93.52 \\
    + ASR~\cite{ASR} & 26.00 & - & 76.87 & - \\
    \rowcolor{shadecolor}+ DAS~(ours) & 26.90 & 4.39 & \textbf{78.04} & \textbf{94.00}\\\rowcolor{white}

    \midrule

    ResNet-101~\cite{he2016deep} & \textbf{44.46} & \textbf{7.85} & 77.35 & 93.56 \\
    + SENet~\cite{SENet} & 49.29 & 7.86 & 77.62 & 93.93 \\
    + BAM~\cite{BAM} & 44.91 & 7.93 & 77.56 & 93.71 \\
    + CBAM~\cite{CBAM} & 49.33 & 7.86 & 78.49 & 94.31 \\
    + SRM~\cite{lee2019srm} & 44.68 & \textbf{7.85} & 78.47 & 94.20 \\
    + Triplet Attention~\cite{TA} & 44.56 & 7.95 & 78.03 & 93.85\\
    + ASR~\cite{ASR} & 45.00 & - & 78.18 & - \\
    \rowcolor{shadecolor}+ DAS~(ours) & 45.89 & 8.12 & \textbf{78.62} & \textbf{94.43}\\\rowcolor{white}

    \midrule

    MobileNetV2~\cite{sandler2018mobilenetv2} & \textbf{3.51} & \textbf{0.32} & 71.88 & 90.29\\
    + SENet~\cite{SENet} & 3.53 & 0.32 & 72.42 & 90.67 \\
    + CBAM~\cite{CBAM} & 3.54 & 0.32 & 69.33 & 89.33 \\
    + Triplet Attention~\cite{TA} & \textbf{3.51} & \textbf{0.32} & 72.62 & 90.77 \\
    \rowcolor{shadecolor} + DAS~(ours) & 3.57 & 0.35 & \textbf{72.79} & \textbf{90.87} \\

   \bottomrule
  \end{tabular}
  
  \end{tabular}
    % \caption{Performance evaluation of image classification models on the ImageNet1k dataset: comparison of top-1, top-5 accuracies and computation. Our model achieved the best accuracies among ResNet-18, ResNet-50 and MobileNetV2 baselines and several attention based models.}
    \caption{Evaluation of image classification models on ImageNet1k dataset, comparing top-1, top-5 accuracies, and computational efficiency. DAS outperforms ResNet-18, ResNet-50, ResNet-101, MobileNetV2, and various other attention-based models, achieving the best accuracies, with only a small increase in parameters and FLOPs.}

  \label{tab:imagenet}
\end{table*}

\subsection{Deformable Attention Gate}

The compressed feature data from the previous step (Eq. \ref{eq1}) represents the feature context that is then passed through a deformable convolution which instead of a regular grid, uses a dynamic grid by an offset of $\Delta p$ introduced in \cite{dai2017deformable, zhu2019deformable}, which as we know helps focus on pertinent image regions. Eq. \ref{eq2} shows the operation of the Deformable Convolution kernel where $K$ is the size of the kernel and its weights are $w_k$ applied on the fixed reference points of $p_{ref}$ the same way as regular kernels in CNNs. $\Delta p$ is a trainable parameter that helps the kernel to find the most relevant features even if they are outside the kernel of the reference. $w_p$ is also another trainable parameter between 0 and 1. Values of $\Delta p$ and $w_p$ are dependent on the features that the kernel is applied on. 
\begin{equation} \label{eq2}
deform(p) = \sum_{k=1}^{K} w_k \cdot w_p \cdot \mathbf{X}(p_{ref,k} + \Delta p_k)
\end{equation}

Following the deformable convolution, we apply Layer Normalization, and then a Sigmoid activation function $\sigma$ (Eq. \ref{eq3}). This convolution operation changes the number of channels from $\alpha \times c$ to the original input $c$.
\begin{equation} \label{eq3}
\mathbf{A} = \sigma(\text{LayerNorm}(deform(\mathbf{X}_c)))
\end{equation}

The output from Eq.~\ref{eq3} represents the attention gate. This gate controls the flow of information from the feature maps, with each element in the gate tensor having values between 0 and 1. These values determine which parts of the feature maps are emphasized or filtered out.
Lastly, to incorporate the DAS attention mechanism into the CNN model, we perform a pointwise multiplication between the original input tensor and the attention tensor obtained in the previous step. 
\begin{equation} \label{eq4}
\mathbf{X_{\text{out}}} = \mathbf{X} \odot \mathbf{A}
\end{equation}

The result of the multiplication in Eq.~\ref{eq4} is the input for the next layer of the CNN model, seamlessly integrating the attention mechanism, without any need to change the backbone architecture.

% \textbf{DAS attention vs deformable attention \cite{zhu2020deformable}.} Previous deformable attentions are mostly designed for transformers not CNNs. For example, \cite{zhu2020deformable} is using a fully connected network (fc) to calculate offsets. While DAS is using $ 3 \times 3$ kernel which is more suitable for CNNs.  \cite{zhu2020deformable} uses deformable attention only for query feature which justify the use of an fc layer while DAS looks at the image features holistically. Our DAS mechanism is also completely a separate module and there is no need to change the main architecture which make it more plugable attention compared to transformer's deformable attentions.

% \setlength{\tabcolsep}{10pt}
\begin{table*}[t!]
  \centering
  \begin{tabular}{ll}
  \begin{tabular}{l cccccccc}
    \toprule
    \multirow{2}{*}{Backbone} &  \multicolumn{6}{c}{Faster R-CNN on MS COCO (\%)} \\
     & Parameters (M) & AP & AP$_{50}$ & AP$_{75}$ & AP$_{S}$ & AP$_{M}$ & AP$_{L}$ \\
    \midrule

    ResNet-50~\cite{he2016deep} &  \textbf{41.7} & 36.4  & 58.4 & 39.1 & 21.5 & 40.0 & 46.6 \\
    ResNet-101~\cite{he2016deep} & 60.6 & 38.5  & 60.3 & 41.6 & 22.3 & 43.0 & 49.8 \\
    \midrule
    SENet-50~\cite{SENet} & 44.2 & 37.7 & 60.1 & 40.9 & 22.9& 41.9 & 48.2 \\
    CBAM-50~\cite{CBAM} & 44.2 & 39.3 & 60.8 & 42.8 & \textbf{24.1} & 43.0 & 49.8 \\
    Triplet Attention-50~\cite{TA} & \textbf{41.7} & 39.3 & 60.8 & 42.7 & 23.4 & 42.8 & 50.3 \\
    \rowcolor{shadecolor} DAS-50~(ours) & 43.0  & \textbf{39.7} & \textbf{60.9} & \textbf{43.2} & 22.8 & \textbf{43.9} & \textbf{51.9} \\

   \bottomrule
  \end{tabular}
  
  \end{tabular}
    % \caption{Performance comparison of different models on MS COCO validation set using Faster R-CNN for object detection task: our model outperforms other attention models and ResNet-101.}
    \caption{Model performance comparison on MS COCO validation using Faster R-CNN for object detection. DAS surpasses other attention models and ResNet-101.}
  \label{tab:coco}
\end{table*}

\paragraph{Comparison of DAS attention and Deformable Attention \cite{zhu2020deformable}} 
Previous deformable attention mechanism, designed primarily for transformers, \cite{zhu2020deformable} employs a fully connected network (FC) to compute offsets, which may not be optimal for CNNs. In contrast, DAS attention utilizes a $3 \times 3$ kernel, better suited for CNNs. While \cite{zhu2020deformable} applies deformable attention exclusively to query features, DAS attention considers image features holistically. Our attention mechanism operates as a separate module without necessitating changes to the main architecture, enhancing its plug-and-play capability over the transformer-based deformable attention approaches.

% \begin{figure}[htp]
%     \centering
%     \includegraphics[width=4cm]{sections/GridSample_algo.jpg}
%     \caption{Our attention module}
%     \label{fig:GS}
% \end{figure}

% \begin{figure}[htp]
%     \centering
%     \includegraphics[height=4cm]{sections/resnet_GS.jpg}
%     \caption{Attention module is added after the skip connections in ResNet models. (4 attention layers)}
%     \label{fig:GS_resnet}
% \end{figure}

%% file: sections/experiments.tex
\section{Experiments}\label{sec:experiments}
% In this section, we evaluate the proposed attention method and compare it with previously proposed attention mechanisms on several challenging computer vision tasks to demonstrate the performance and efficiency of our method.
\vspace{-3pt}
\paragraph{Training Setup}

For image classification, we used CIFAR100~\cite{krizhevsky2009learning}, Stanford Dogs~\cite{khosla2011novel}, and ImageNet1k~\cite{deng2009imagenet} datasets, and for object detection, MS COCO~\cite{lin2014microsoft}. We employed ResNet~\cite{he2016deep}  and MobileNetV2~\cite{sandler2018mobilenetv2} architectures as per \cite{TA}.

For ImageNet experiments, we adopted settings from \cite{TA}: ResNet training with batch size 256, initial LR 0.1, and weight decay 1e-4 for 100 epochs. LR scaled at 30$^{\text{th}}$, 60$^{\text{th}}$, and 90$^{\text{th}}$ epochs by a factor of 0.1. MobileNetV2: batch size 96, initial LR 0.045, weight decay 4e-5, LR scaled by $0.98^{epoch}$.

For CIFAR100 and Stanford Dogs datasets, we compared with Triplet Attention~\cite{TA} and Vanilla Resnet. We conducted a hyperparameter search for ResNet-18, and used the same setup for all of the baselines: 300 epochs, batch size 128, initial LR 0.1, weight decay 5e-4, LR decay at 70$^{\text{th}}$, 130$^{\text{th}}$, 200$^{\text{th}}$, 260$^{\text{th}}$ by a scale factor of 0.2. For Stanford Dogs: batch size 32, LR 0.1, weight decay 1e-4, CosineAnnealing LR scheduler, random flip and crop for image pre-processing.

For object detection, we used Faster R-CNN on MS COCO with MMdetection toolbox \cite{chen2019mmdetection}, with batch size 16, initial LR 0.02, weight decay 0.0001, and ImageNet-1k pre-trained backbone. We mitigated noise by initial training of the backbone, training both the backbone and the rest of the model for a few epochs. The weights obtained from this initial training served as an initialization point for our subsequent training process. We consistently employed the SGD optimizer.

\subsection{Image Classification}

\cref{tab:cifar_dogs} demonstrates that the addition of Triplet Attention \cite{TA} slightly improves the accuracy of ResNet-18 CIFAR100 (0.3\%) but decreases the accuracy by 1.36\% on the Stanford Dogs dataset. However, DAS improves the accuracy of ResNet-18 by 0.79\% and 4.91\% on CIFAR100 and Stanford Dogs, respectively. Similar to ResNet-18, the addition of Triplet attention \cite{TA} to ResNet-50 has a negative impact on the backbone model for Stanford Dogs, while DAS enhances the backbone model by 2.8\% and 4.47\% on CIFAR100 and Stanford Dogs, respectively, showing DAS's performance consistency across small and large models. 

% In our experiments, we observed a remarkable result where our proposed DAS18 method consistently outperformed not only the base ResNet-18 model but also deeper architectures, including ResNet-50. The ability of DAS18 to outperform deeper models for not very large-datasets has important implications for practical applications where computational resources and model complexity are critical considerations. By demonstrating that DAS18 can achieve higher accuracy with a shallower architecture, we open up new possibilities for deploying efficient and effective deep-learning models in resource-constrained environments.

Interestingly, we observed that our proposed DAS-18 method outperformed not only the base ResNet-18 model but also deeper architectures on CIFAR100 and Stanford Dogs datasets, including ResNet-50, while using 2.26G less FLOPs. This makes DAS-18 a compelling option for mobile applications.

Results for ImageNet classification are presented in \cref{tab:imagenet}. When the DAS attention gate is applied to ResNet-18, it demonstrates remarkable improvements in classification accuracy. The DAS results in a top-1 accuracy of 72.03\% and a top-5 accuracy of 90.70\%. This outperforms other existing methods such as SENet \cite{SENet}, BAM \cite{BAM}, CBAM \cite{CBAM}, Triplet Attention \cite{TA}, and EMCA \cite{EMCA} showcasing the efficacy of DAS in enhancing model performance.

DAS with a depth of 50 achieves a top-1 accuracy of 78.04\% and a top-5 accuracy of 94.00\%. It achieves the best performance while using 32\% less FLOPs and 1.39M less parameters compared to the second best performer (GSoP-Net \cite{GSoP-Net}). ResNet-50 + DAS attention also outperforms ResNet-101 in terms of top-1 accuracy, with 0.69\% more accuracy at $\sim$60\% of FLOPs and number of parameters. ResNet-101 + DAS attention achieves the best top-1 accuracy (78.62\%) compared to other attention modules with less parameters compared to SENet \cite{SENet} and CBAM \cite{CBAM}.

\begin{table}[t!]
% \small
  \centering
  \begin{tabular}{ll}
  \begin{tabular}{l cc}
    \toprule
    \multirow{2}{*}{Method} &   CIFAR100 & Stanford Dogs \\
    & (Acc \%) & (Acc \%)\\
    \midrule

    ResNet-18~\cite{he2016deep} & 78.25 & 61.50\\
    + Triplet Attention~\cite{TA} & 78.55 & 60.14 \\
    + DAS~(ours) & \textbf{79.04} & \textbf{66.41} \\

    \midrule

    ResNet-50~\cite{he2016deep} & 77.74 & 62.58 \\
    + Triplet Attention~\cite{TA} & 79.22 & 60.55\\
    \rowcolor{shadecolor} + DAS~(ours) & \textbf{80.54} & \textbf{67.05}\\

   \bottomrule
  \end{tabular}
  
  \end{tabular}
    % \caption{Performance comparison (accuracy \%) of different architectures on the CIFAR100 and Stanford Dogs datasets, The proposed method achieves the best result.}
    \caption{Performance (\%) on CIFAR100 and Stanford Dogs datasets, with our method DAS, achieving the highest accuracy.}
  \label{tab:cifar_dogs}
\end{table}

% including SENet \cite{SENet}, BAM \cite{BAM}, CBAM \cite{CBAM}, GSoP-Net \cite{GSoP-Net}, A2-Nets \cite{A2-Nets}, GCNet \cite{GC-Net}, GALA \cite{GALA}, SRM \cite{lee2019srm}, and Triplet Attention \cite{TA} with a lower number of parameters comparing to SENet, CBAM, GSoP-Net, A2-Net, GCNet, GALA and ABN. This highlights the substantial benefit of DAS in improving the classification capabilities of deeper networks.

On the lightweight MobileNetV2, DAS maintains its effectiveness. It achieves a top-1 accuracy of 72.79\% and a top-5 accuracy of 90.87\%, outperforming SENet \cite{SENet}, CBAM \cite{CBAM}, and Triplet Attention \cite{TA}, while being computationally efficient with a low FLOP count of 0.35G.

\subsection{Object Detection}

\cref{tab:coco} shows results from our object detection experiments using the Faster R-CNN model on the challenging MS COCO dataset. The metrics used for evaluation include average precision (AP), AP at different intersections over union (IoU) thresholds (AP$_{50}$, AP$_{75}$), and class-specific AP for small (AP$_{S}$), medium (AP$_{M}$), and large (AP$_{L}$) objects.

The choice of backbone architecture significantly impacts object detection performance. In our evaluation, ResNet-50, ResNet-101, SENet-50, CBAM-50, and Triplet Attention-50 serve as strong baselines. Our DAS-50 model surpasses all other backbones in terms of AP, AP$_{50}$, AP$_{75}$, AP$_{M}$, and AP$_{L}$ scores, with a lower number of parameters compared to ResNet-101, SENet-50 and CBAM-50.

\subsection{Design Evolution and Ablation Studies}\label{sec:ablation}

% Before we proposed the mentioned method, we tested various pixel-wise attention ideas. The two of them that led us to the current idea are shown in Fig.\ref{fig:ideas} (a) and (b) and the result on Stanford Dogs are presented in \cref{tab:ideas}.

% \textbf{(a)}: involved concatenating the input with a DAS of itself, followed by a convolutional layer that combined the input and learned distant pixels. While this approach demonstrated promise, it achieved an accuracy of 65.00\% on the Stanford Dogs dataset. DAS is an operation that takes an input tensor and a grid tensor and computes the spatial interpolation of near-neighbor pixels around the given grid.

\begin{figure}[t]

    \centering
    \resizebox{0.13\textwidth}{!}{
        \begin{subfigure}{1.2\linewidth}
            \centering
            \includegraphics{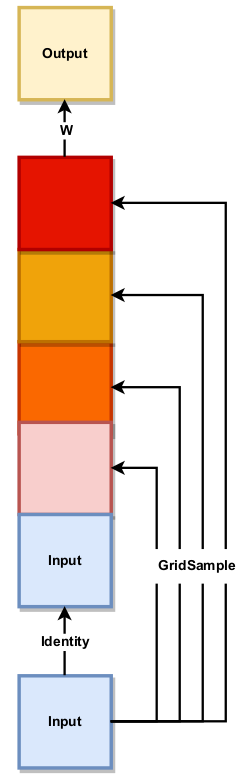}
            \caption*{\Huge (a)}
            \label{fig:idea1}
        \end{subfigure}
    }
    \resizebox{0.15\textwidth}{!}{
        \begin{subfigure}{1.2\linewidth}
            \centering
            \includegraphics{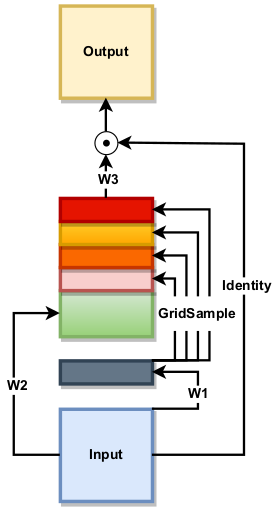}
            \caption*{\Huge (b)}
            \label{fig:idea2}
        \end{subfigure}
    }
    \resizebox{0.15\textwidth}{!}{
        \begin{subfigure}{1.2\linewidth}
            \centering
            \includegraphics{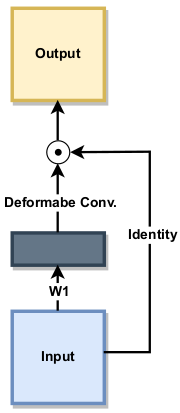}
            \caption*{\Huge (c): proposed method}
            \label{fig:idea3}
        \end{subfigure}
    }

    \resizebox{0.09\textwidth}{!}{
        \begin{subfigure}{0.7\linewidth}
            \centering
            \includegraphics{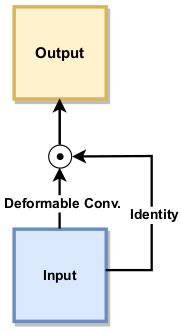}
            \caption*{\Huge (d)}
        \end{subfigure}
    }
    \resizebox{0.09\textwidth}{!}{
        \begin{subfigure}{0.7\linewidth}
            \centering
            \includegraphics{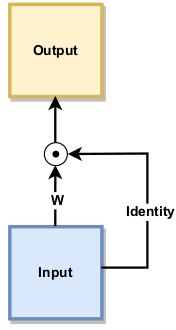}
            \caption*{\Huge (e)}
        \end{subfigure}
    }
    \resizebox{0.09\textwidth}{!}{
        \begin{subfigure}{0.7\linewidth}
            \centering
            \includegraphics{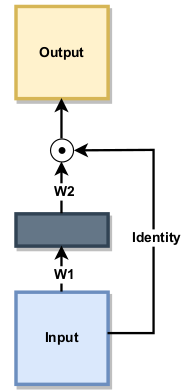}
            \caption*{\Huge (f)}
        \end{subfigure}
    }
    \resizebox{0.09\textwidth}{!}{
        \begin{subfigure}{0.7\linewidth}
            \centering
            \includegraphics{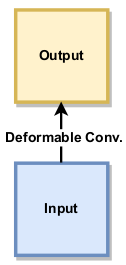}
            \caption*{\Huge (g)}
        \end{subfigure}
    }
    \resizebox{0.09\textwidth}{!}{
        \begin{subfigure}{0.7\linewidth}
            \centering
            \includegraphics{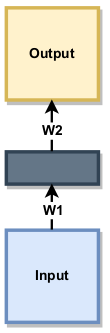}
            \caption*{\Huge (h)}
        \end{subfigure}
    }

% \caption{(a and b): Ablation studies on different ideas: (a) Concatenation of a feature tensor with its deformed grids following a convolution layer to capture long range dependencies. (b) Similar to (a) but with compressed channels resulting in less FLOPs and parameters. (c) The proposed method: compressing the channels of input and using a deformable convolutional layer to generate attention weights. (d) to (h): Ablation study on each component of (c) by removing or changing some parts explained in Sec. \ref{sec:ablation}. We added them between each main blocks of ResNet after the skip connection. Results presented in Table \ref{tab:ideas} shows the superiority of (c) in terms of accuracy and computation cost.}

\caption{(a and b): Ablation studies on ideas in Sec. \ref{sec:ablation}: (a) Concatenating a feature tensor with deformed grids, followed by convolution for global dependencies. (b) Similar to (a) with compressed channels for reduced FLOPs and parameters. (c) Our method: channel compression and deformable convolution for attention to relevant information. (d) to (h): Ablation on each component of (c) explained in Sec. \ref{sec:ablation}. Table \ref{tab:ideas} demonstrates (c)'s superior accuracy and computational efficiency. }
    \label{fig:ideas}
\end{figure}

Before finalizing the design of DAS, we explored two pixel-wise attention concepts. These are depicted in Fig. \ref{fig:ideas} (a) and (b), with corresponding results on the Stanford Dogs dataset in Table \ref{tab:ideas}.

\textbf{(a)}: We concatenated the input with a GridSample of itself, followed by a convolutional layer that integrated both the input and information from distant pixels. While this approach showed potential, it achieved an accuracy of 65.00\% on the Stanford Dogs dataset. GridSample is a differentiable PyTorch feature that interpolates neighboring pixels spatially based on a given grid tensor.

% Formulas:

% $output = DAS(x)$

% $output = concatenation(x, output)$

% $output = conv(output)$

% \textbf{(b)}: It was a variation of the first idea, employing compressed inputs and DAS outputs while calculating weights to suppress additional information in the features. This approach showed a slight improvement compared to the first idea, reaching an accuracy of 65.21\% with a lower computation cost.

\textbf{(b)}: We extended the initial concept by using compressed inputs and GridSample outputs to compute weights for suppressing extraneous information in the features. This refinement yielded a modest improvement over the first idea, achieving an accuracy of 65.21\% while reducing computational overhead.

% Formulas:

% $compressed\_input = conv1(x)$

% $Compressed\_DAS = DAS(conv2(x))$

% $output = concatenation(compressed\_input, Compressed\_DAS)$

% $output = conv3(output)$

% $output = torch.mul(x, output)$
% \begin{table}[t!]
% % \small
%  \centering
% \begin{tabular}{cccc}
% \multicolumn{1}{l}{Idea} & \multicolumn{1}{l}{Accuracy(Dogs)} & \multicolumn{1}{l}{FLOPs (B)} & Parameters (M) \\
% Idea \#1                 & 65                                  & 1.99                         & 12.766            \\
% Idea \#2                 & 65.21                               & 1.846                         & 11.801          \\
% Current method           & \textbf{66.41}                      & 1.856                         & 11.817          
% \end{tabular}
% \end{table}

% \setlength{\tabcolsep}{10pt}

% \subsection{Image Classification Results}

\begin{table}[t!]
  \centering
  \begin{tabular}{ll}
  \begin{tabular}{l ccc}
    \toprule
    \multirow{1}{*}{Methods in Fig.\ref{fig:ideas}} &  \multicolumn{3}{c}{Stanford Dogs} \\
     ResNet-18 + & Parameters & FLOPs & Top-1 (\%)\\
    \midrule

    a & 12.766M  & 1.99G & 65.00 \\
    b & 11.801M & 1.846G & 65.21 \\
    \rowcolor{shadecolor} c (ours) & \textbf{11.817M} & \textbf{1.856G} & \textbf{66.410}\\

    \rowcolor{white}d & 11.923M & 1.915G & 65.338 \\
    e & 12.048M & 1.879G & 65.291  \\
    f & 11.843M & 1.849G & 66.107 \\
    g & 11.925M & 1.917G & 60.851 \\
    h & 11.819M & 1.858G & 61.375 \\

   \bottomrule
  \end{tabular}
  
  \end{tabular}
    % \caption{Ablation study on different ideas and components of the proposed method presented in Fig.\ref{fig:ideas}: (a and b) are similar ideas, (d to h) are studies on each component of the proposed method (c). Assessing model efficiency and performance on the Stanford Dogs dataset shows that all components of our method has a positive impact on its performance and efficiency.}
    \caption{Ablation study of DAS components in Fig. \ref{fig:ideas} and explained in Sec. \ref{sec:ablation}: (a, b) Design evolution analyses, (d-h) Component analyses of the proposed method (c). Evaluation on the Stanford Dogs dataset reveals the positive influence of each component on model performance and efficiency.}
  \label{tab:ideas}
\end{table}

\begin{table}[]
\centering
\begin{tabular}{cccc|cccc|c}
\toprule
\multicolumn{4}{c}{\textbf{First Norm. Layer}} & \multicolumn{4}{c}{\textbf{Second Norm. Layer}} & \multirow{2}{*}{Top-1(\%)} \\
BN & FN & IN & LN & BN & FN & IN & LN\\
\midrule
% BN & FN & IN & LN & BN & FN & IN & LN & Accuracy\\
  \checkmark   &    &    &    &  \checkmark  &    &     &  &    65.24   \\
     &  \checkmark  &    &    &    &  \checkmark  &     &  &    64.64   \\
     &    &  \checkmark  &    &    &    &  \checkmark   &  &    66.27   \\
     &    &    &  \checkmark  &    &    &     &  \checkmark &   64.66    \\
   \checkmark  &    &    &    &    &    &     & \checkmark &    65.78   \\

     &  \checkmark  &    &    &    &    &     & \checkmark &    65.41   \\
     &    &    &  \checkmark  &    &    &     & \checkmark &    65.15   \\
     &    &  \checkmark  &    &  \checkmark  &    &     &  &    65.94   \\
     &    &  \checkmark  &    &    &  \checkmark  &     &  &    65.25   \\
     % &    & \checkmark   &    &    &    &  \checkmark   &  &    66.25  \\
     &    & \checkmark   &    &    &    &     &  &   66.21  \\
      \rowcolor{shadecolor} &    & \checkmark   &    &    &    &     & \checkmark &    \textbf{66.41}   \\

\bottomrule
\end{tabular}
% \caption{Ablation study on the nomralization layers in our proposed attention gate. We have tested batchnorm (BN), simple feature norm (FN), InstanceNorm (IN), GroupNorm with number of groups of 1 GN(1), GN with number of groups as input channels GN(c) in ResNet-18 + our attention on Stanford Dogs. Our proposed method is the last line with the highest accuracy.}
\caption{Ablation study of normalization layers in our attention gate. We evaluated BatchNorm (BN), Simple Feature Norm (FN), IntsanceNorm (IN), and LayerNorm (LN) in ResNet-18 + our DAS attention on Stanford Dogs. Our method (last line) achieved the best accuracy.}
\label{tab:normalization_layers}
\end{table}

To evaluate our design decisions \textbf{(c)} we conducted various ablation studies:

\textbf{(d)} Removing the initial part and relying solely on deformable convolution led to reduced accuracy (65.338\%), emphasizing the importance of the first convolution layer.

\textbf{(e)} Removing deformable convolution while keeping the initial part increased computation and decreased accuracy (65.291\%), indicating the need for multiple layers for precise attention modeling.

\textbf{(f)} Replacing deformable convolution with depthwise separable convolutions improved accuracy (66.107\%), but it was still outperformed by our method, highlighting the advantage of deformable convolution in focusing attention on relevant information.

\textbf{(g)} Excluding attention modules and only using deformable convolution drastically decreased accuracy, emphasizing the significance of attention behavior.

\textbf{(h)} Similarly, excluding attention modules and using additional layers showed low accuracy, emphasizing the preference for using these layers as an attention module.

Our attention method \textbf{(c)} outperformed all configurations, achieving the best accuracy (66.410\%). This underscores the effectiveness of our context-aware attention mechanism in focusing attention on relevant features even outside of kernel boundaries and enhancing model performance. Table \ref{tab:normalization_layers} demonstrates the effect of different normalization layers on the attention module.

In summary, our experiments demonstrate our method's superiority in accuracy and computational efficiency compared to other ideas and configurations, establishing it as a valuable addition to pixel-wise attention modeling.

\begin{figure}
    \centering
    \begin{tikzpicture}
        \begin{axis}[
            width=0.8\columnwidth,  % Adjust the width to fit within one column
            height=0.4\textwidth,
            xlabel={FLOPs (G)},
            ylabel={Top-1 (\%)},
            legend pos=north west,
            legend columns=2,
            ymin=65,
            ymax=68,
            xmin=1.8,
            xmax=2.0,
            xtick={1.8,1.9,2},
            xticklabels={1.8,1.9,2.0},
            ytick={65,66,67,68},
            ymajorgrids=true,
            grid style=dashed,
            nodes near coords style={
                anchor=south, % Adjust the anchor position
                font=\scriptsize, % Adjust the font size
                rotate=90, % Rotate the labels if needed
                inner sep=1pt, % Adjust the distance between the point and the label
            },
        ]
        
        \addplot[mark=*,nodes near coords={$\alpha=0.01$}] coordinates {
            (1.829, 65.315)
        };
        
        \addplot[mark=*,nodes near coords={$\alpha=0.1$}] coordinates {
            (1.842, 66.224)
        };
        
        \addplot[mark=*,nodes near coords={$\alpha=0.2$}] coordinates {
            (1.856, 66.410)
        };

        \addplot[mark=*,nodes near coords={$\alpha=0.5$}] coordinates {
            (1.899, 66.527)
        };
        
        \addplot[mark=*,nodes near coords={$\alpha=1.0$}] coordinates {
            (1.970, 66.830)
        };
        
        % Adding lines between nodes
        % \addplot[quiver={u=0,v=1},-stealth,white] coordinates {(1.842, 66.224) (1.842, 66.224)};
        % \addplot[quiver={u=0,v=1},-stealth,white] coordinates {(1.856, 66.410) (1.856, 66.410)};
        % \addplot[quiver={u=0,v=1},-stealth,white] coordinates {(1.899, 66.527) (1.899, 66.527)};
        
        \end{axis}
    \end{tikzpicture}
    % \caption{Ablation study on the impact of compression coefficient $\alpha$: performance of ResNet18 + our attention on the Stanford Dogs dataset shows that the model is not very sensitive to the only added hyperparameter when $\alpha>0.1$ . Default $\alpha$ for this paper is 0.2.}
    \caption{Ablation study on compression coefficient $\alpha$: ResNet18 + our attention on Stanford Dogs indicates low sensitivity to this added hyperparameter when $\alpha>0.1$. Default $\alpha$ used in our implementation for this paper is 0.2.}
    \label{fig:alphas}
\end{figure}
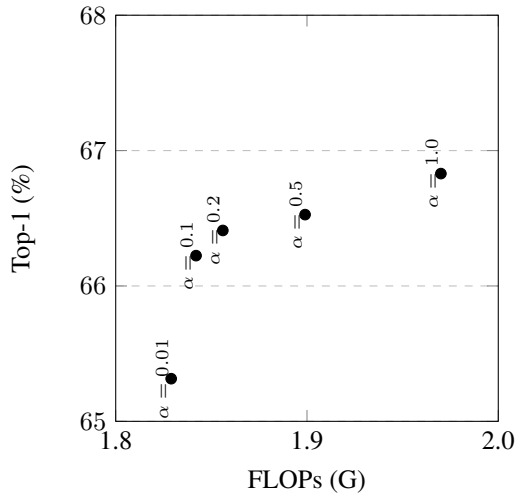

We examined the impact of varying the parameter $\alpha$ from 0.01 to 1. Increasing $\alpha$ increases both FLOPs and parameters. Our findings in Fig. \ref{fig:alphas} indicate that alpha values greater than 0.1 yield favorable results. Typically, there exists a trade-off between FLOPs and accuracy. Consequently, we opted for $\alpha=0.2$ in the majority of our investigations.

% Furthermore, to investigate the effect of the number of attention layers, we added attention layers after all skip connections. Although it will slightly improve the performance, FLOPS, and parameters will be increased, especially in larger models. So we concluded that four attention layers are worth more in terms of computation cost and accuracy. More studies have been done, and the final proposed attention model was chosen with care to be simple, efficient, and accurate on both small and large datasets.

We examined the impact of the number of attention layers. Adding attention layers after all skip connections slightly enhances performance but significantly increases FLOPs and parameters, especially in larger models. Empirically, our observation is that four attention gate layers strike a good balance between computation cost and accuracy. We also conducted studies on attention gate locations, ultimately choosing an attention model that is simple, efficient, and accurate for both small and large datasets.

% \begin{table}[]
% \begin{tabular}{clcc}
% \multicolumn{1}{l}{\begin{tabular}[c]{@{}l@{}}Effect of \# layers\\ (ablation study)\end{tabular}} & \begin{tabular}[c]{@{}l@{}}Accuracy\\ (Dogs)\end{tabular} & \multicolumn{1}{l}{\begin{tabular}[c]{@{}l@{}}FLOPs\\ (B)\end{tabular}} & \begin{tabular}[c]{@{}c@{}}Parameters\\ (M)\end{tabular} \\
% \begin{tabular}[c]{@{}c@{}}GS18\_all\_layers\\ (TEST 300)\end{tabular}                             & \multicolumn{1}{c}{66.783}                                & 1.889                                                                   & 11.944                                                   \\
% \begin{tabular}[c]{@{}c@{}}GS18\_4\_layers\\ (current)\end{tabular}        & \multicolumn{1}{c}{66.41}                                 & 1.856                                                                   & 11.817                                                   \\
% GS50\_all\_layers                                                                                  & \textbf{}                                                 & 5.115                                                                   & 30.477                                                   \\
% \begin{tabular}[c]{@{}c@{}}GS50\_4\_layers\\ (current)\end{tabular}                                &                                                           & 4.387                                                                   & 26.899                                                  
% \end{tabular}
% \end{table}

%% file: sections/visualizations.tex
\subsection{Salient Feature Detection Effectiveness}\label{sec:visualizations}

The objective of applying an attention mechanism in any task is to pay increased attention to relevant features, while at the same time paying less or no attention to irrelevant features. We believe that the performance improvements presented in the earlier sections are primarily due to the effectiveness of our gate in focusing and increasing attention to salient features in the image. In this section, we visualize the extent to which our attention mechanism meets the above objective. For this, we use gradCAM~\cite{selvaraju2017grad}, a function that produces a heatmap showing which parts of an input image were important for a classification decision made by the trained network. The color scheme used in the heatmap is red to blue with blue representing lower importance.

\begin{figure}[t]
    \centering
    \begin{tabular}{cccc}
    % \begin{tabular}{m{2cm} m m m
        \textbf{Image} & \textbf{Method} & \textbf{GradCam} & $\textbf{sfd}$ \\

        \hline
        \multirow{3}{*}{\resizebox{0.10\textwidth}{!}{\includegraphics{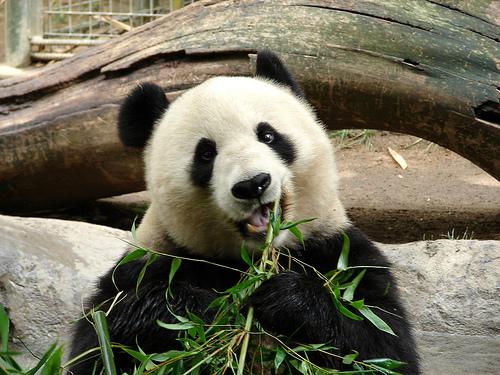}}}
        & ResNet & 
        \resizebox{0.06\textwidth}{!}{\includegraphics{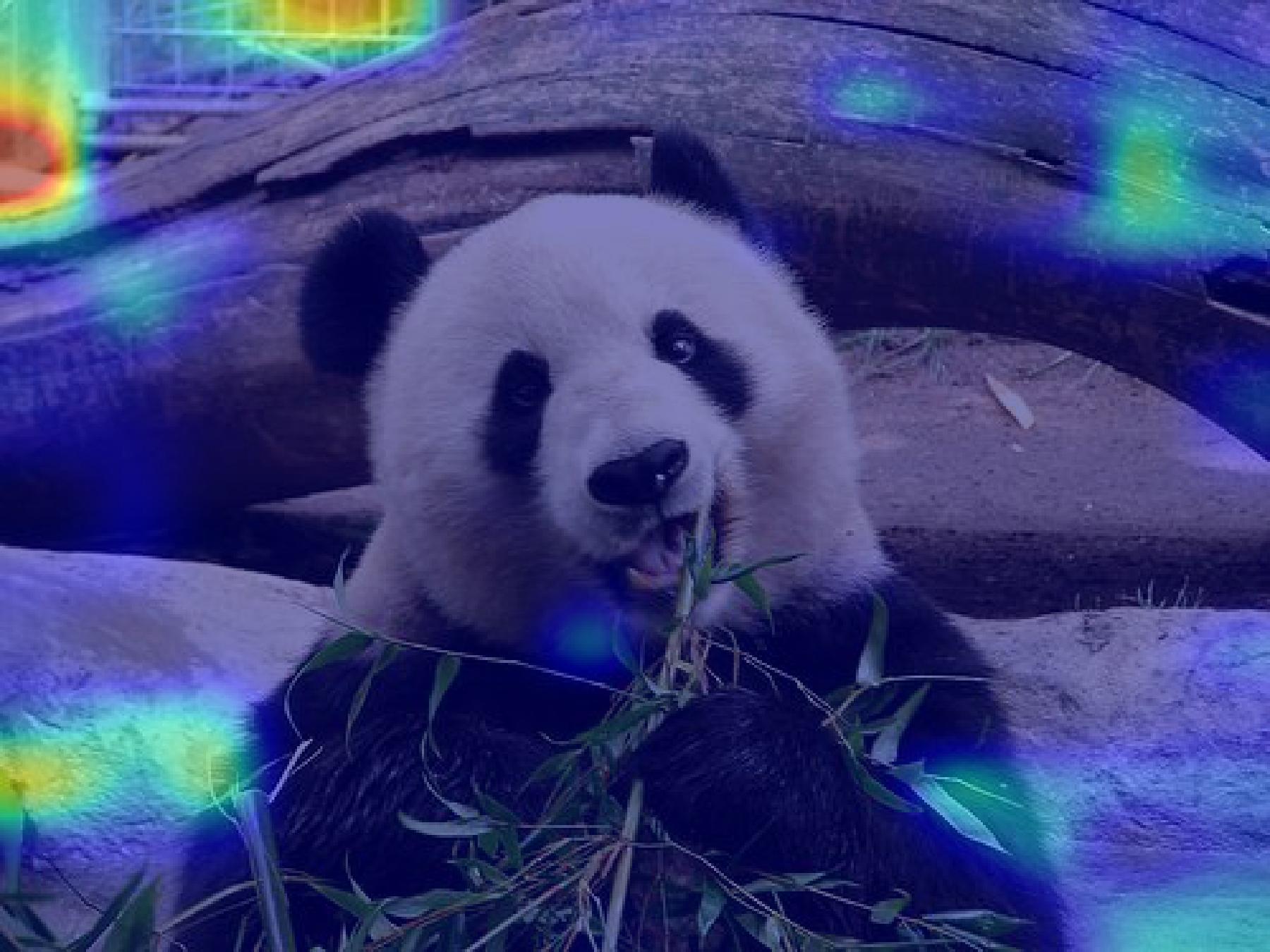}}\hspace{0mm}
        \resizebox{0.06\textwidth}{!}{\includegraphics{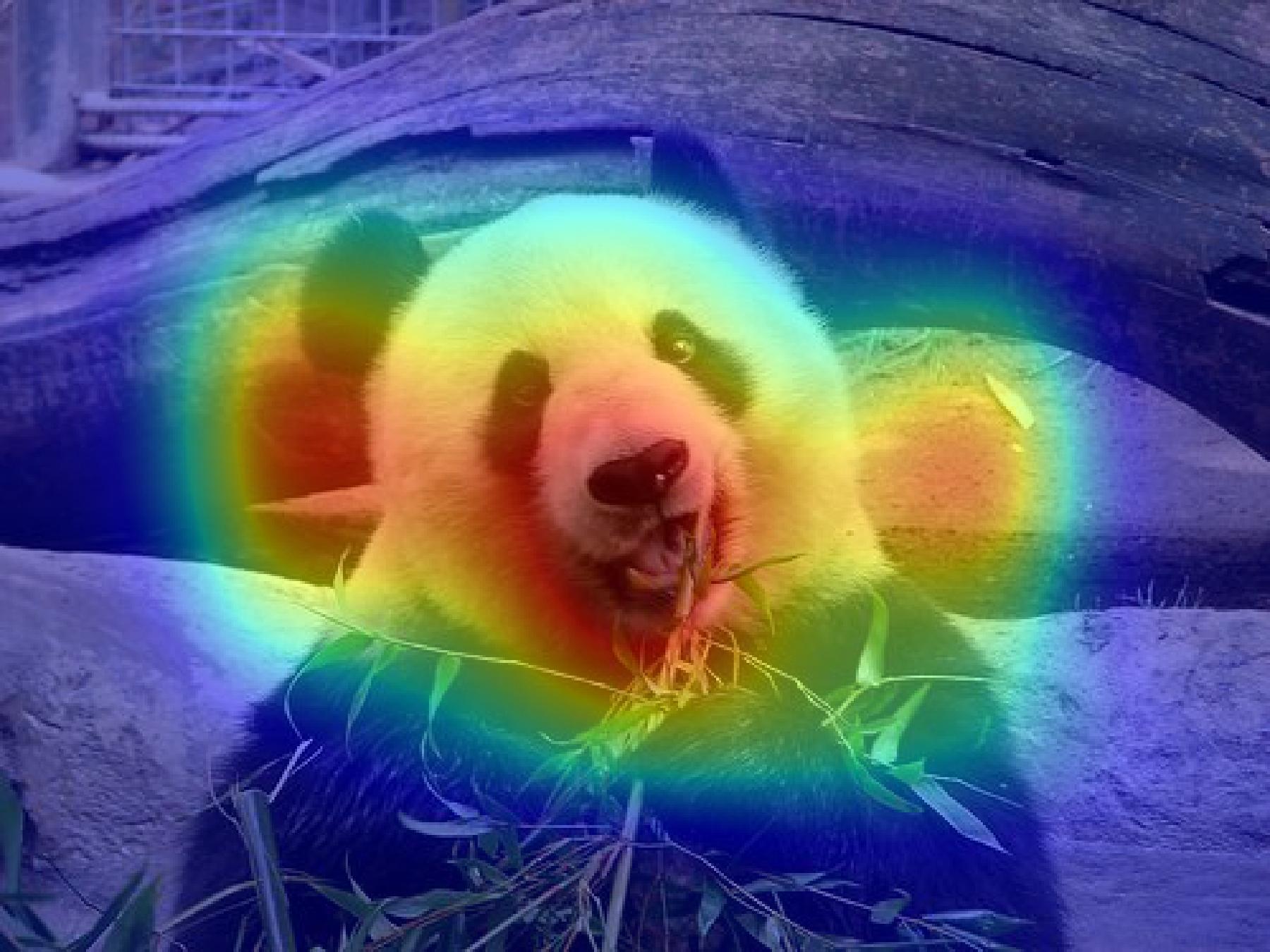}} & 0.47 \\
        \cline{2-4}
        & DAS &
        \resizebox{0.06\textwidth}{!}{\includegraphics{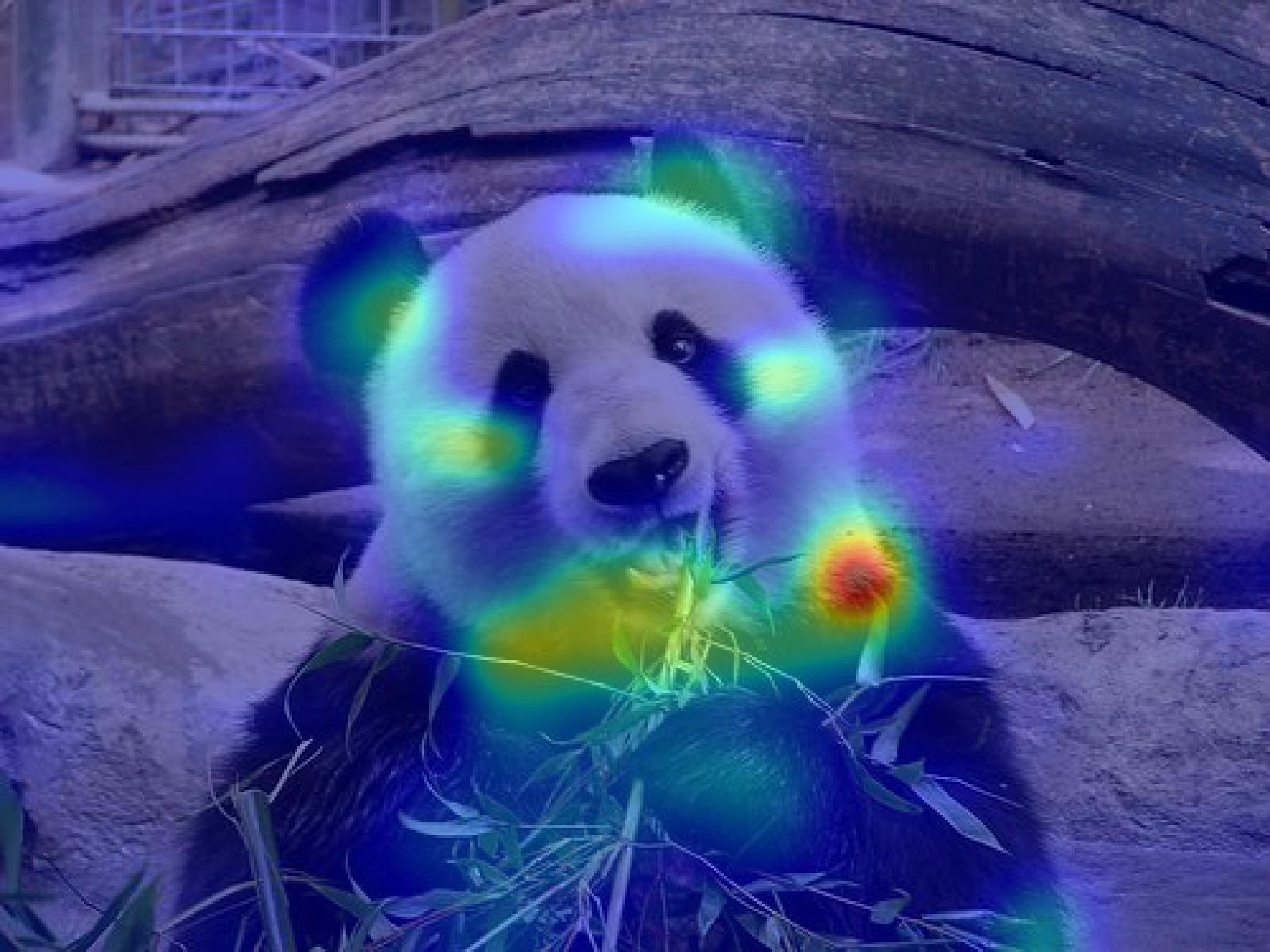}}\hspace{0mm}
        \resizebox{0.06\textwidth}{!}{\includegraphics{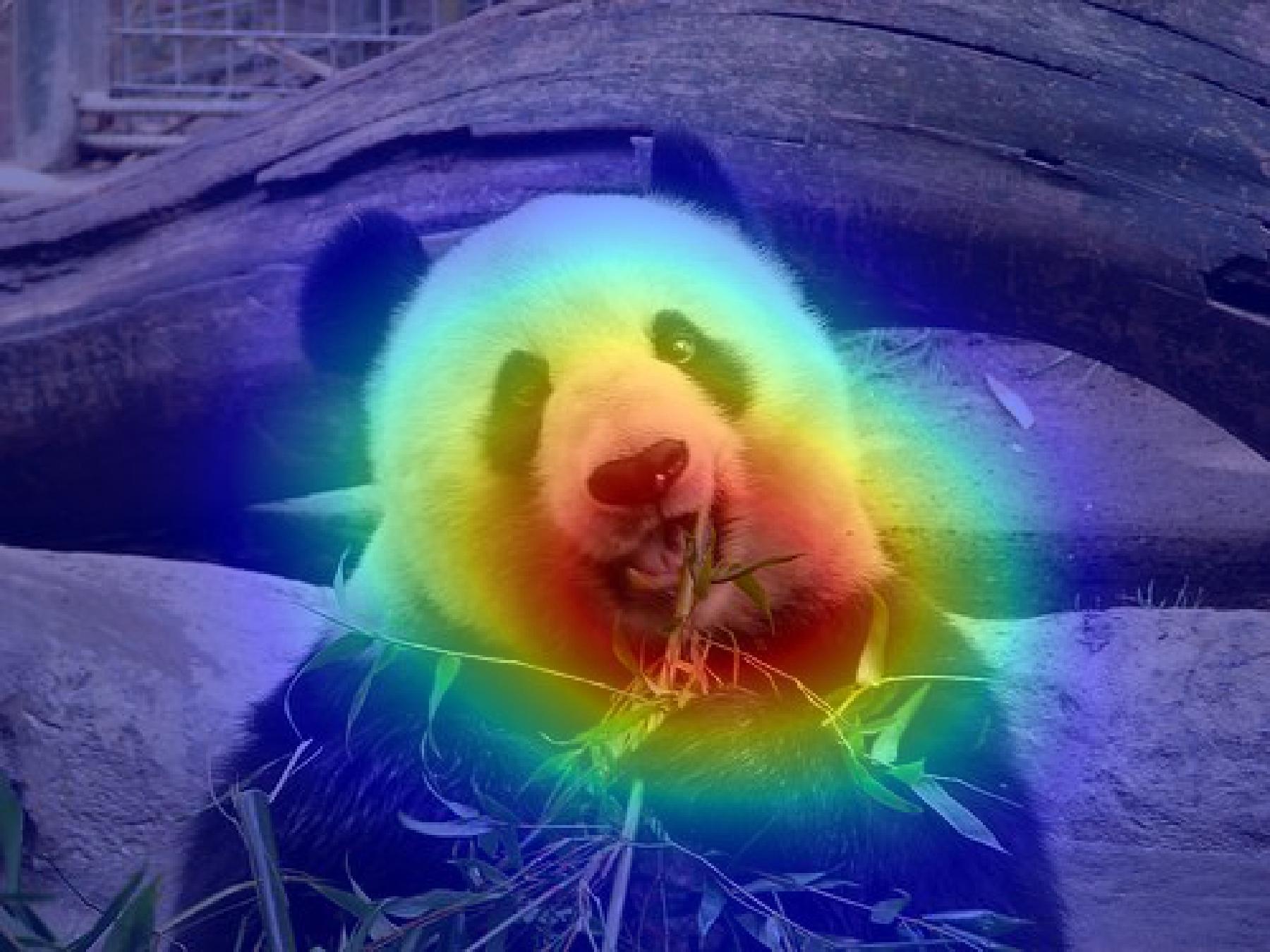}} & \textbf{0.68} \\

        \hline
        
        \multirow{3}{*}{\resizebox{0.095\textwidth}{!}{\includegraphics{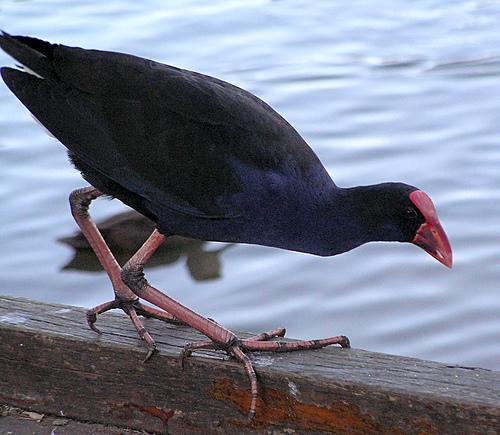}}}
        & ResNet & 
        \resizebox{0.06\textwidth}{!}{\includegraphics{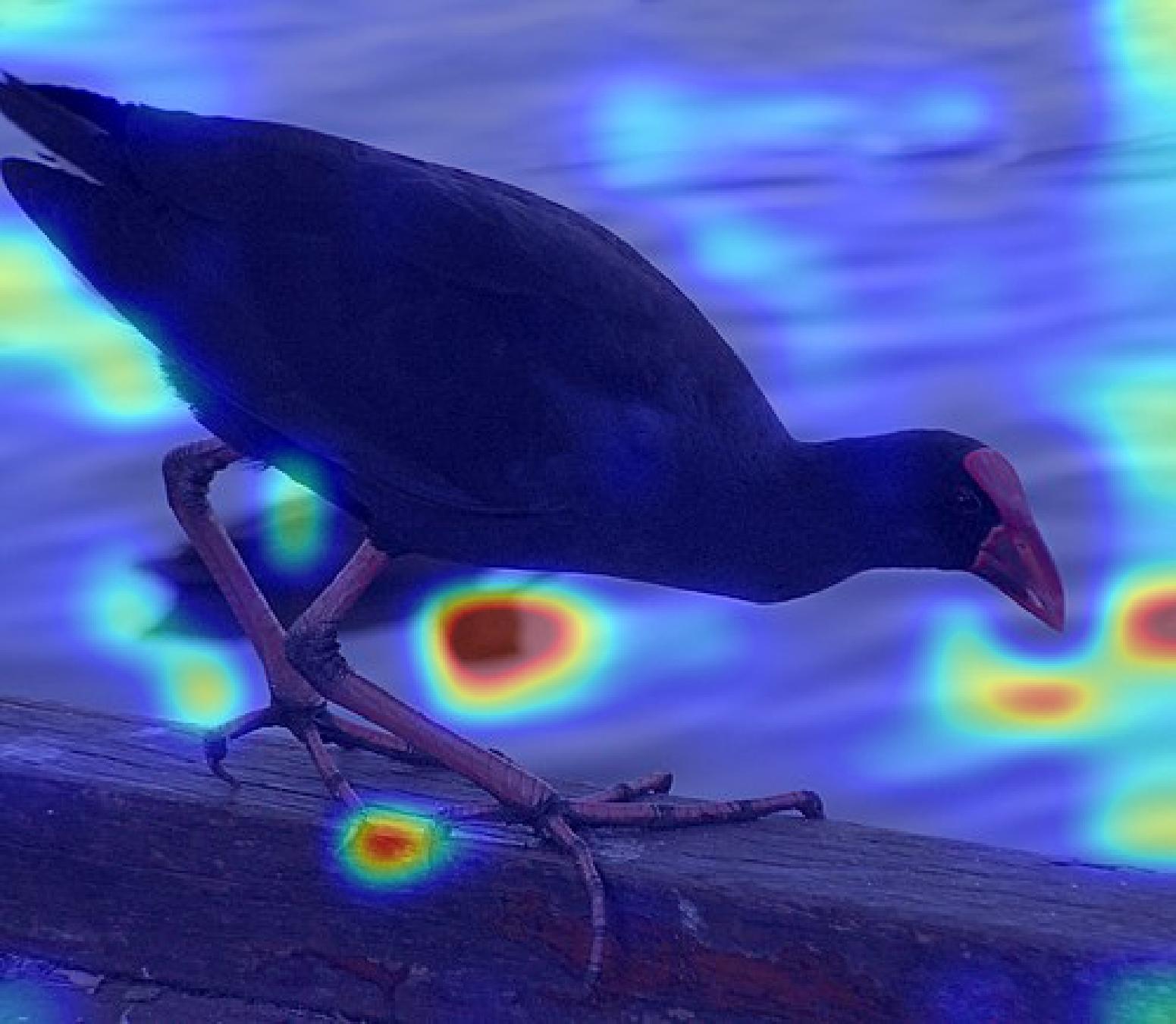}}\hspace{0mm}
        \resizebox{0.06\textwidth}{!}{\includegraphics{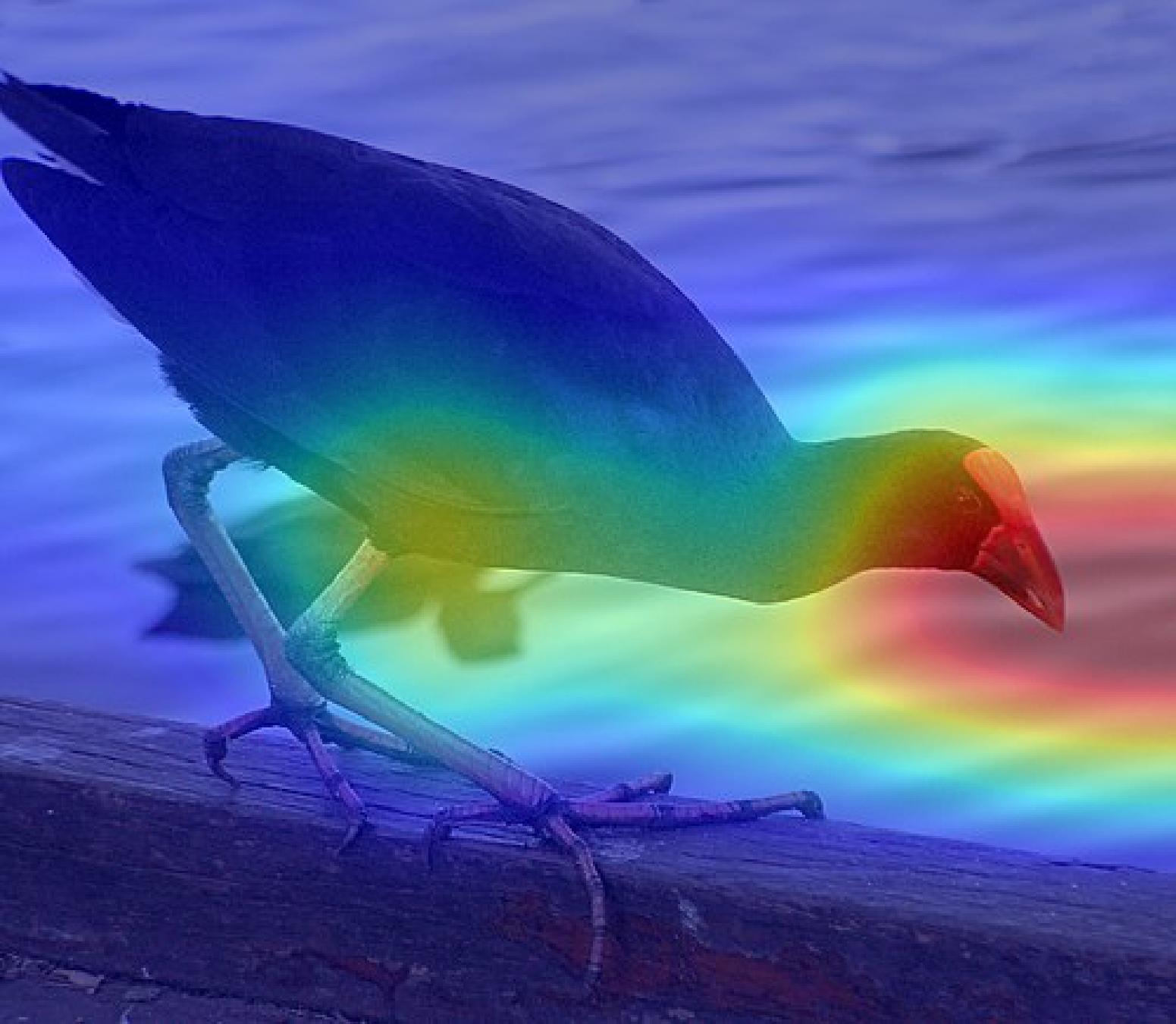}} & 0.15 \\
        \cline{2-4}
        & DAS &
        \resizebox{0.06\textwidth}{!}{\includegraphics{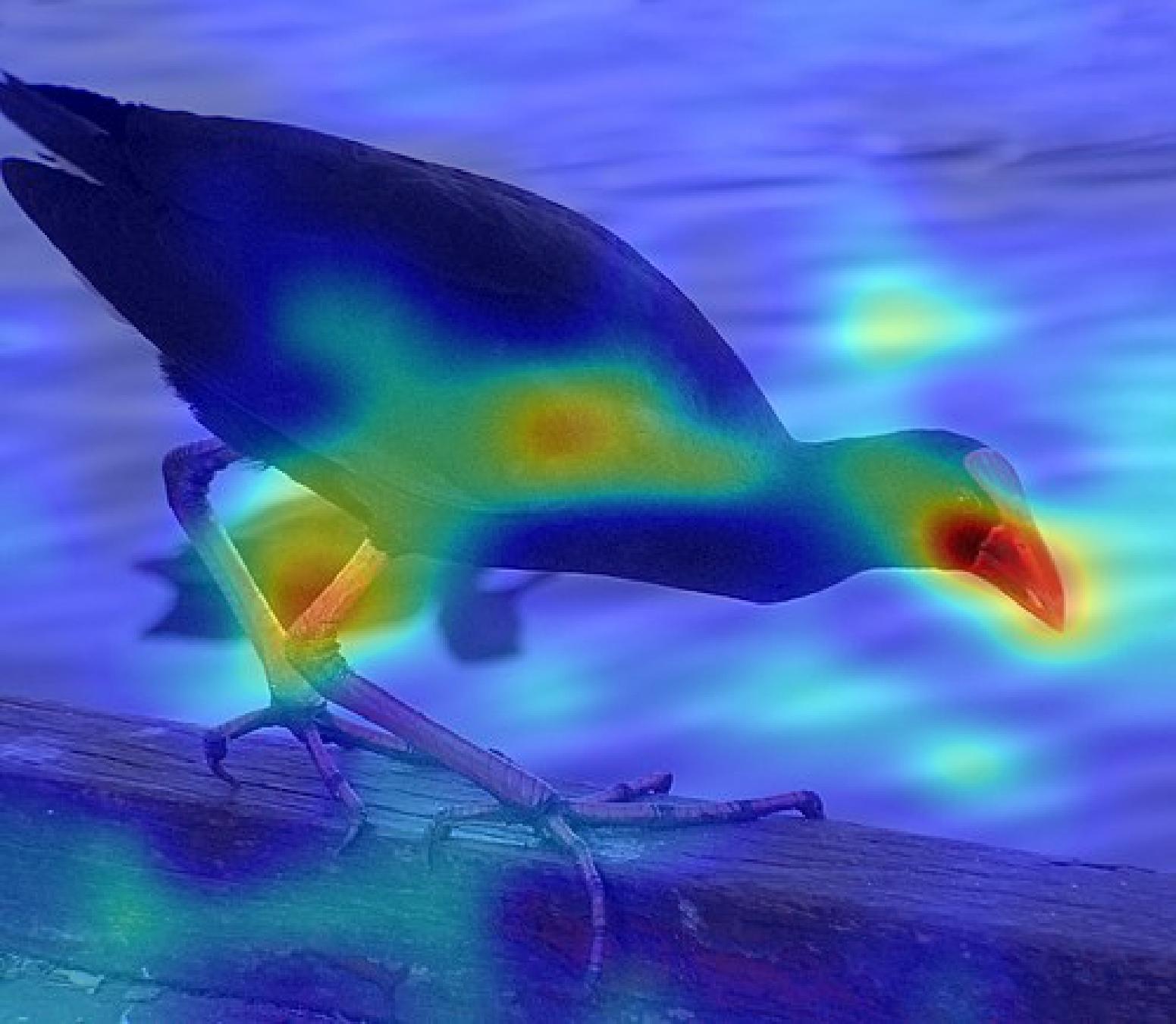}}\hspace{0mm}
        \resizebox{0.06\textwidth}{!}{\includegraphics{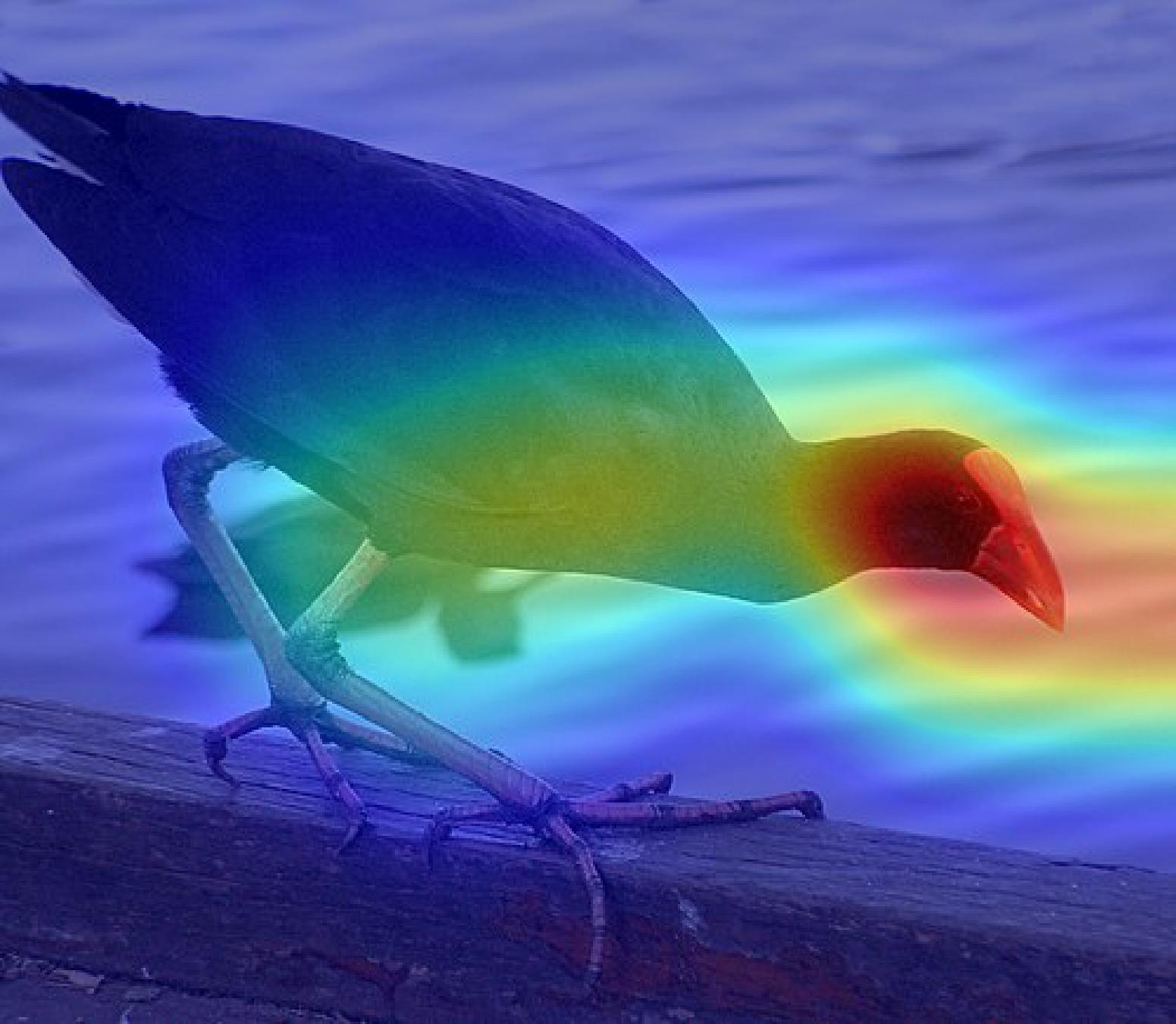}} & \textbf{0.41} \\

        \hline
        
        \multirow{3}{*}{\resizebox{0.10\textwidth}{!}{\includegraphics{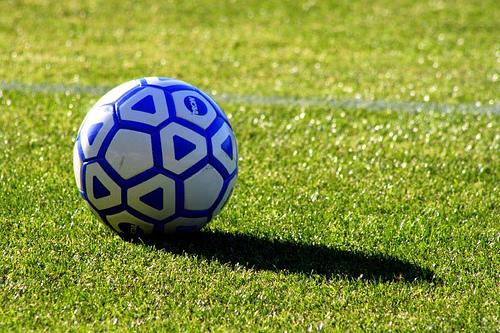}}}
        & ResNet & 
        \resizebox{0.06\textwidth}{!}{\includegraphics{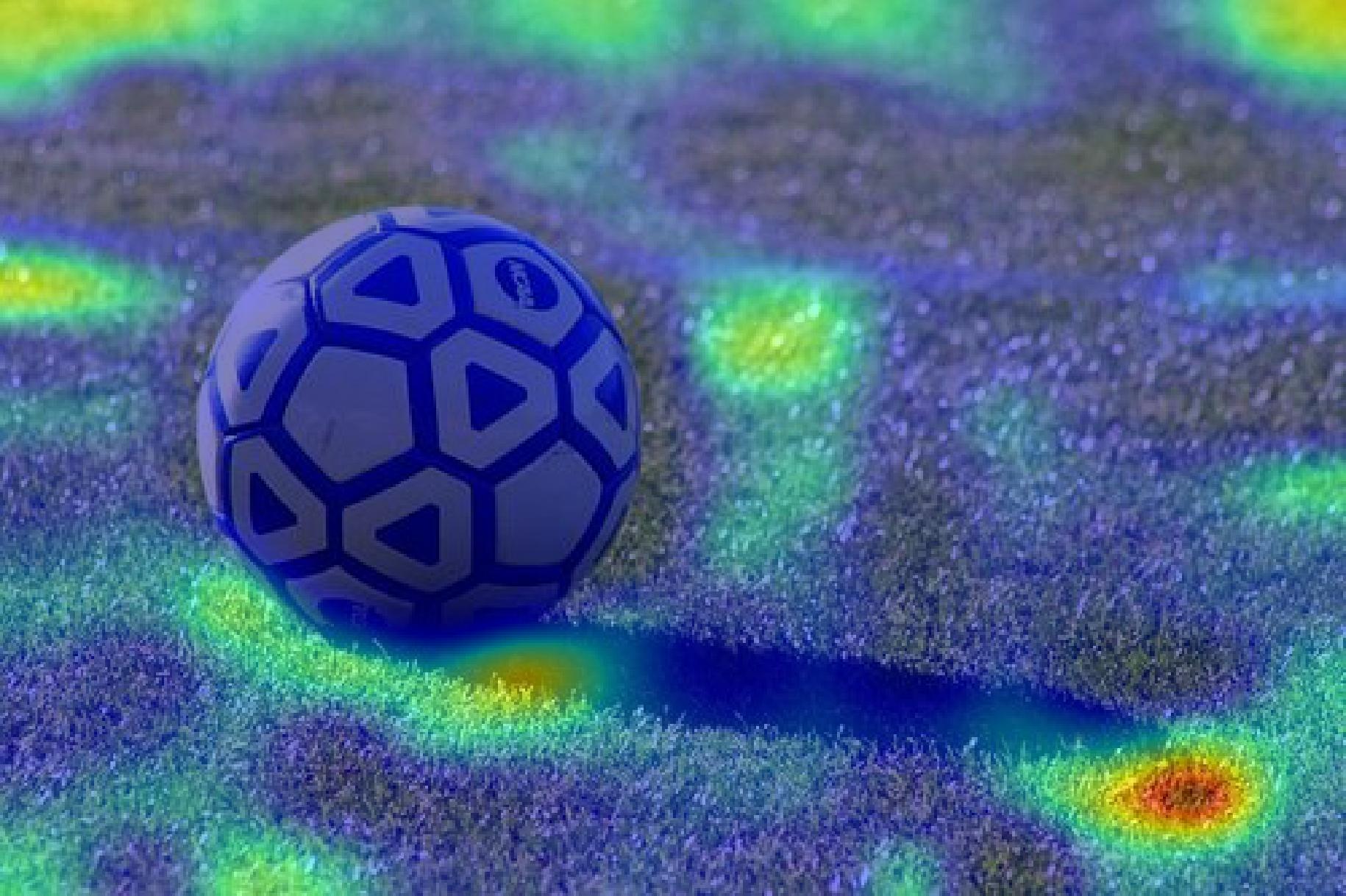}}\hspace{0mm}
        \resizebox{0.06\textwidth}{!}{\includegraphics{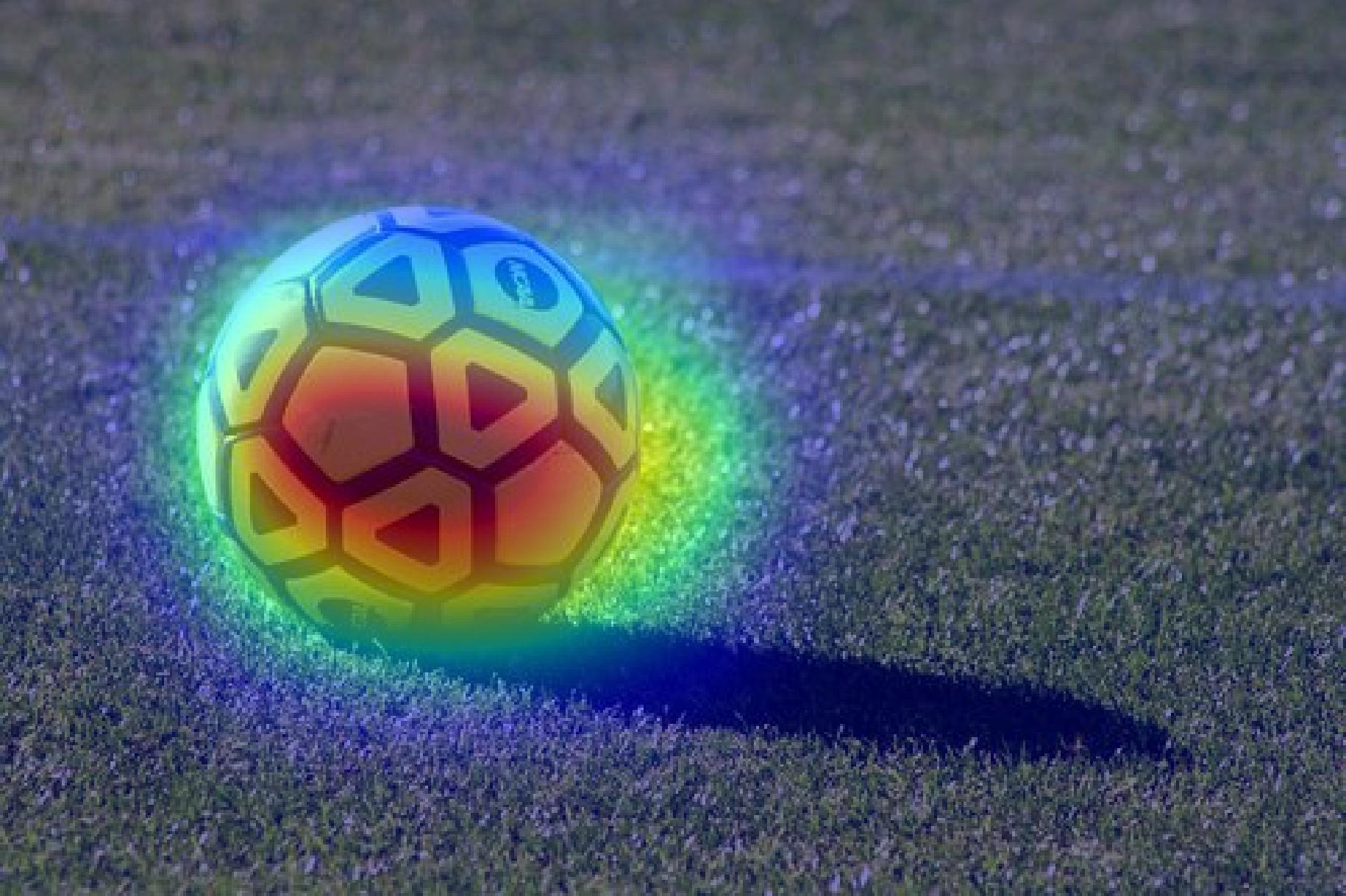}} & 0.87 \\
        \cline{2-4}
        & DAS &
        \resizebox{0.06\textwidth}{!}{\includegraphics{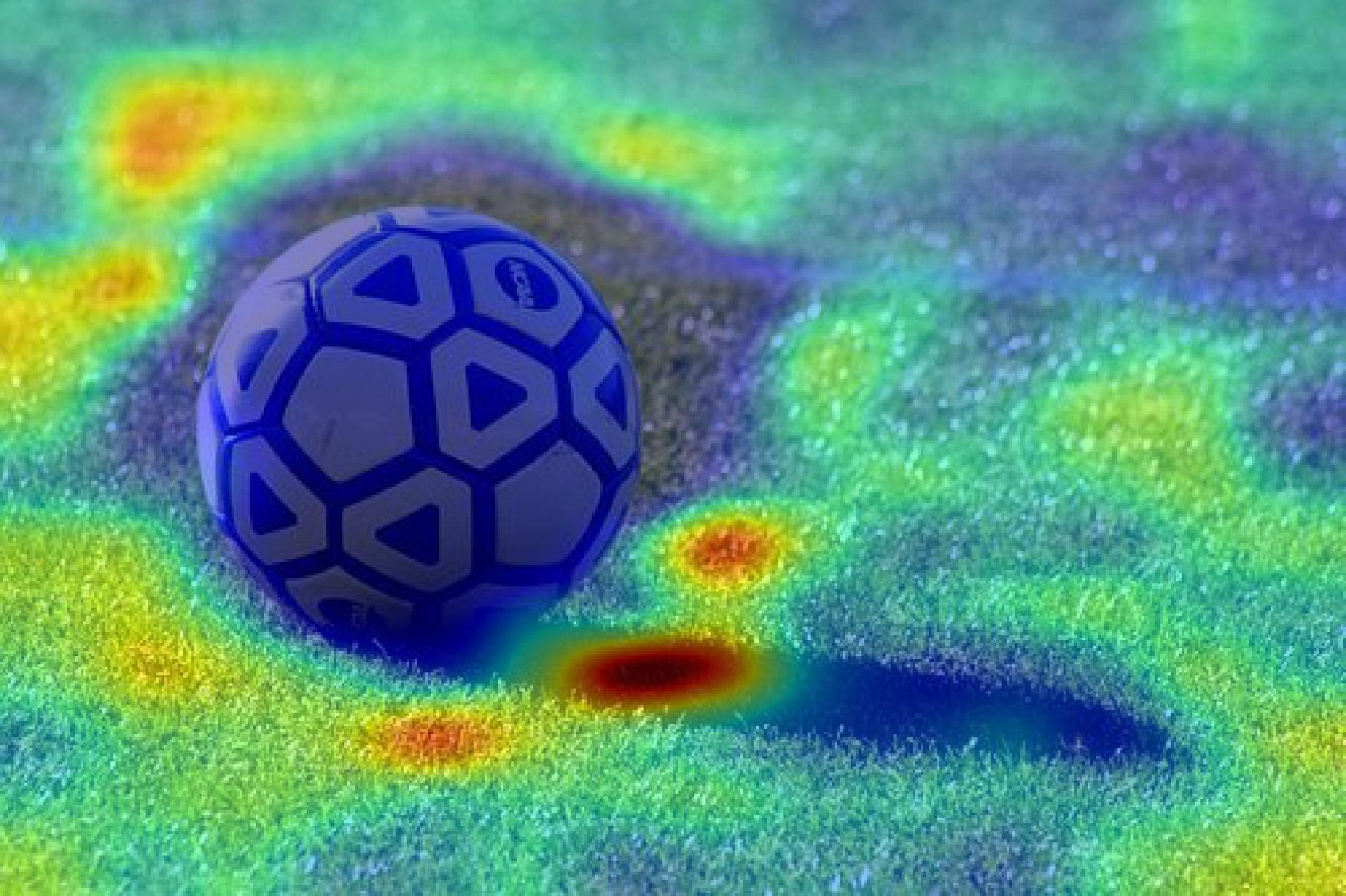}}\hspace{0mm}
        \resizebox{0.06\textwidth}{!}{\includegraphics{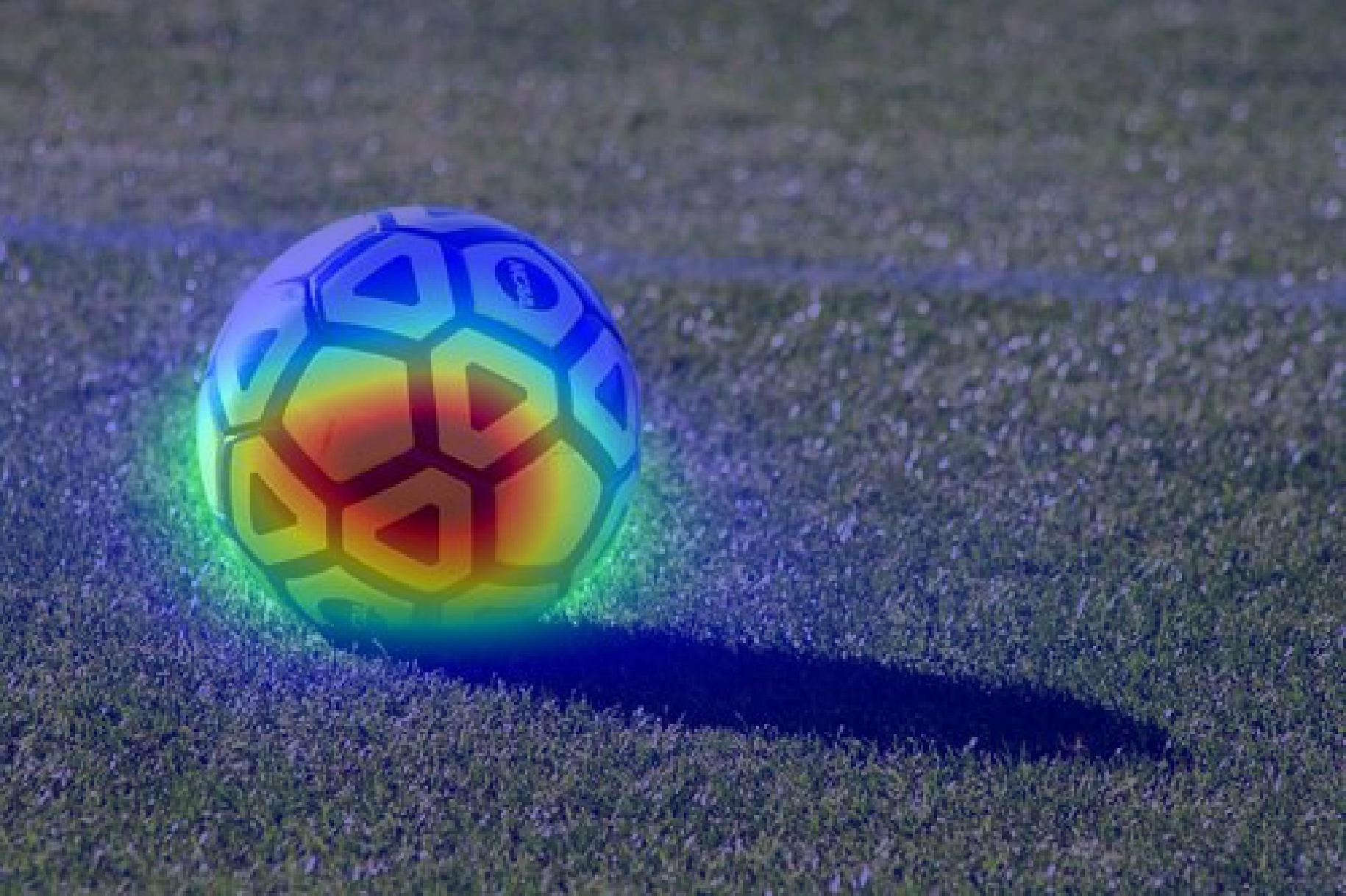}} & \textbf{0.99} \\

         \hline
         
        \multirow{3}{*}{\resizebox{0.10\textwidth}{!}{\includegraphics{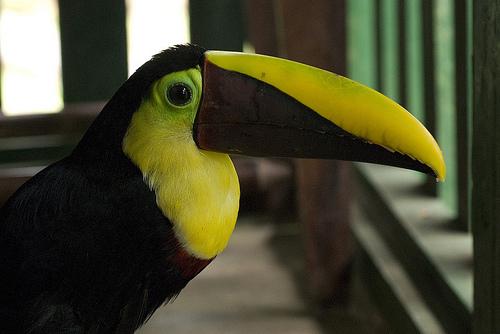}}}
        & ResNet & 
        \resizebox{0.06\textwidth}{!}{\includegraphics{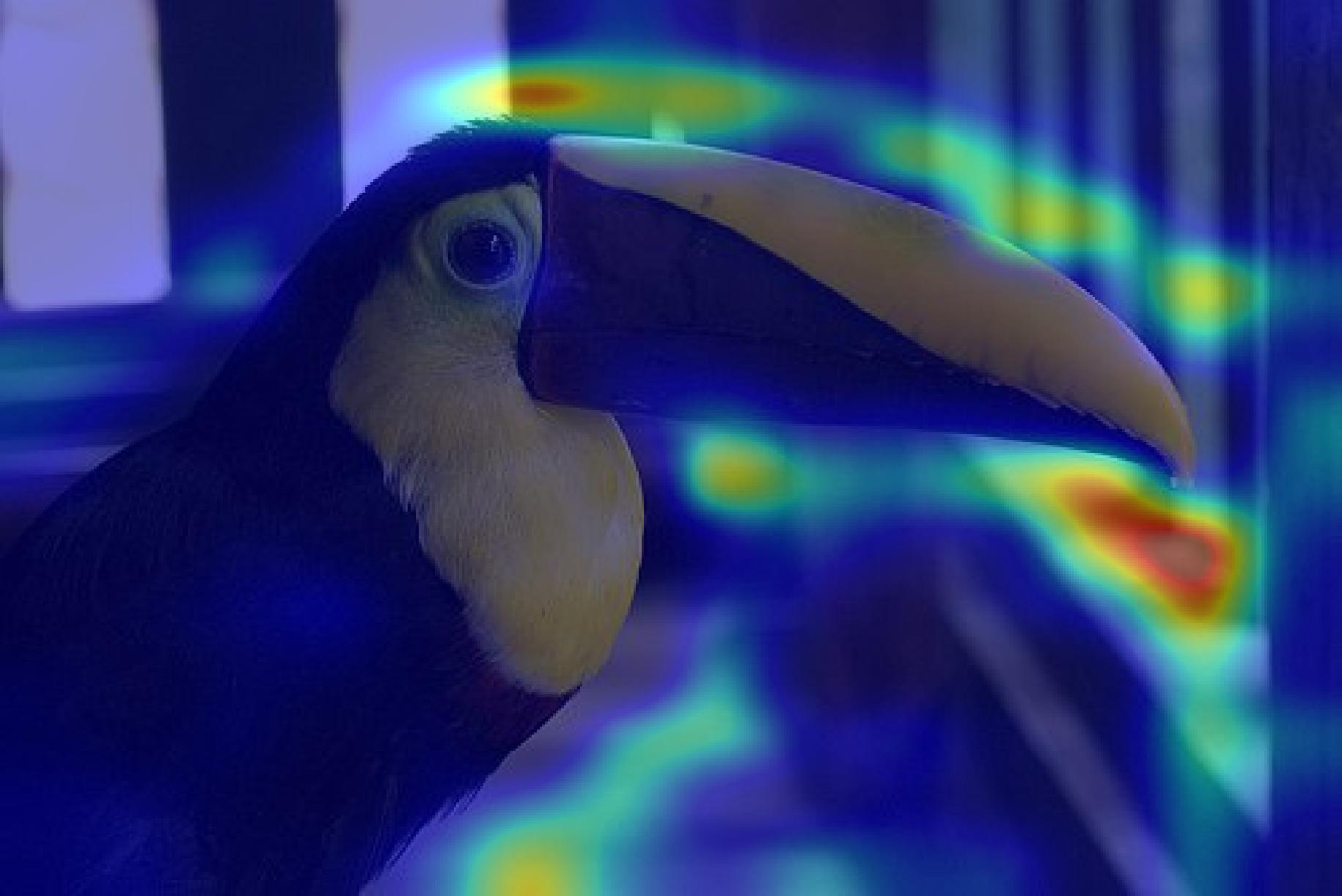}}\hspace{0mm}
        \resizebox{0.06\textwidth}{!}{\includegraphics{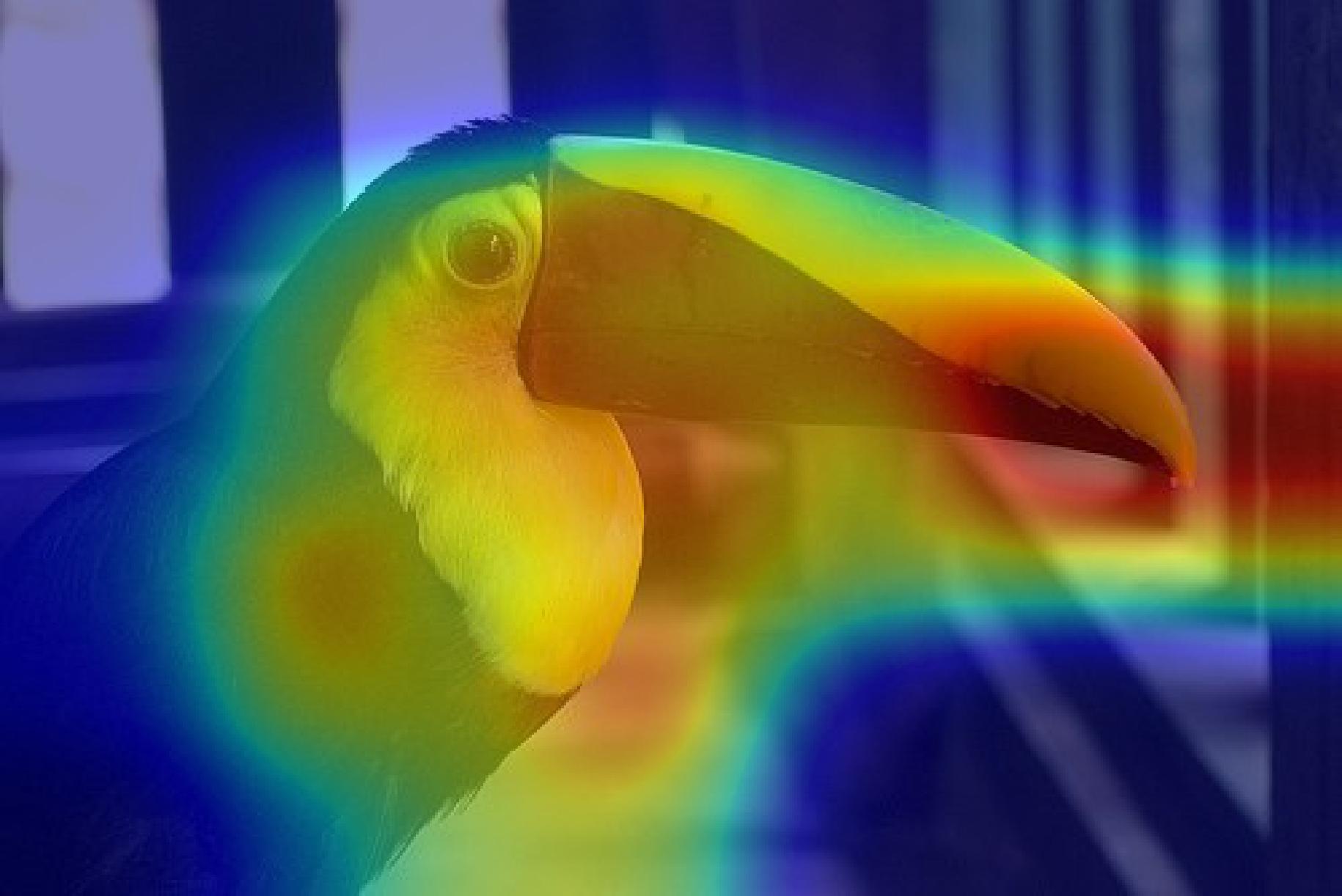}} & 0.38 \\
        \cline{2-4}
        & DAS &
        \resizebox{0.06\textwidth}{!}{\includegraphics{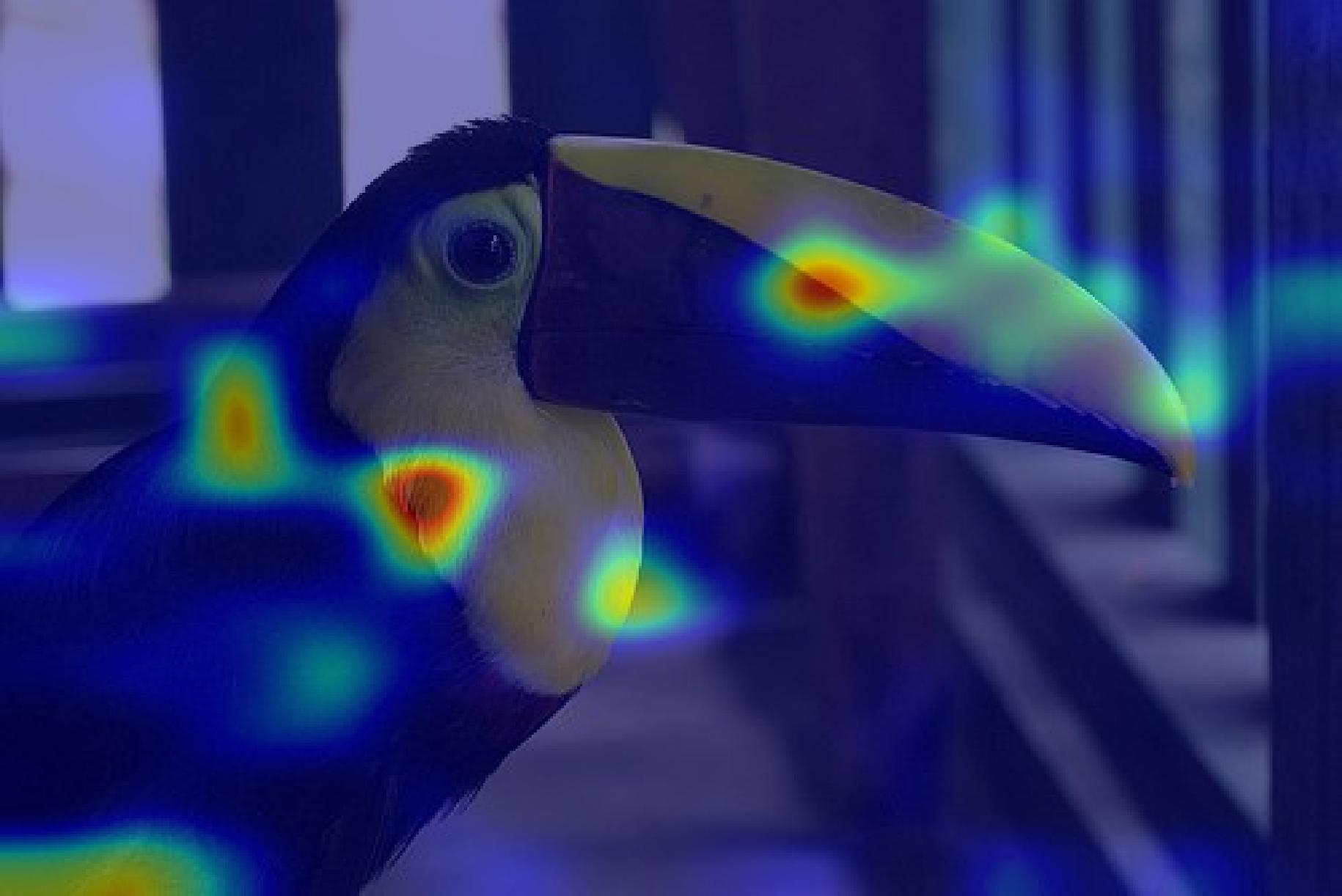}}\hspace{0mm}
        \resizebox{0.06\textwidth}{!}{\includegraphics{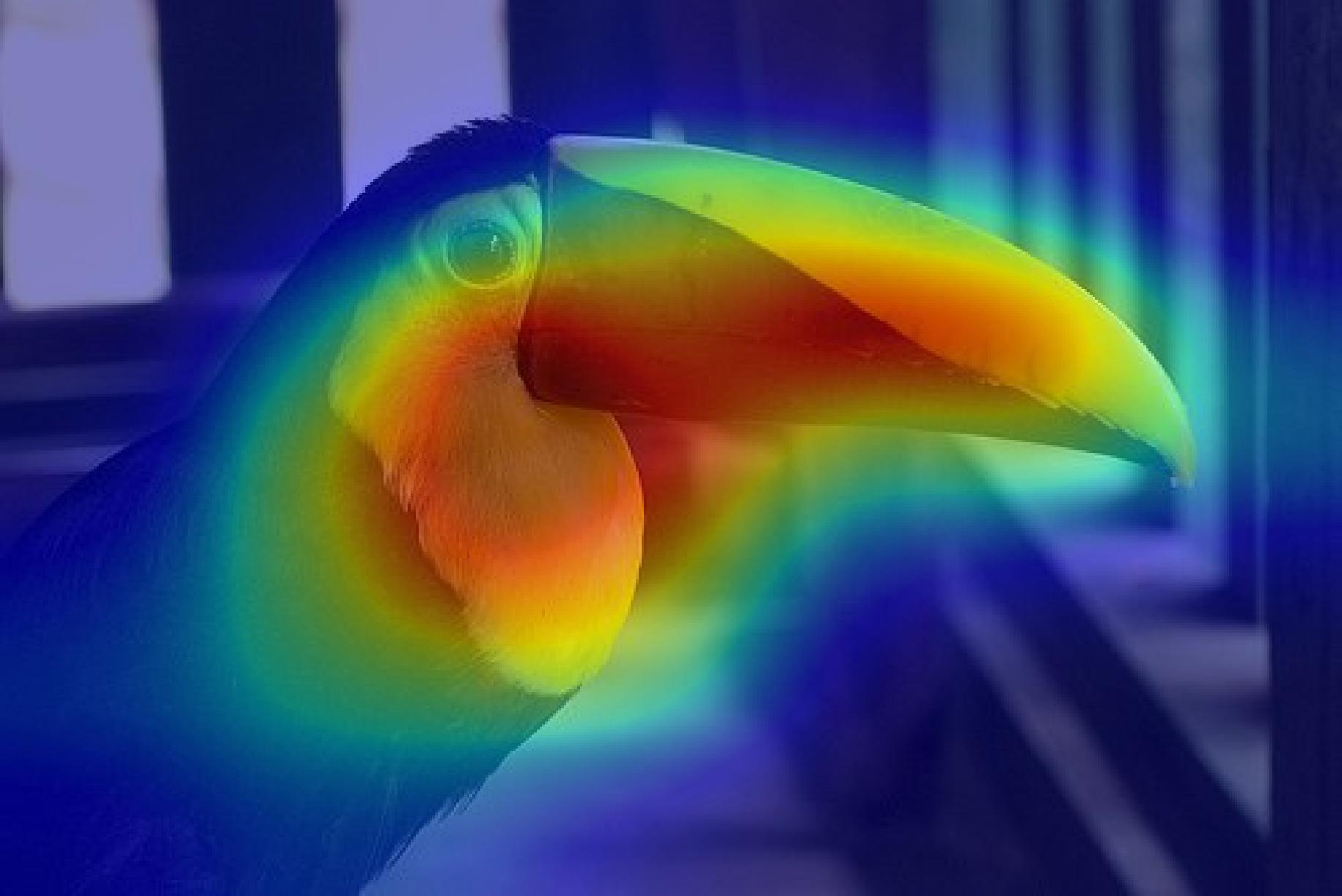}} & \textbf{0.84} \\

        \hline
        
        \multirow{3}{*}{\resizebox{0.10\textwidth}{!}{\includegraphics{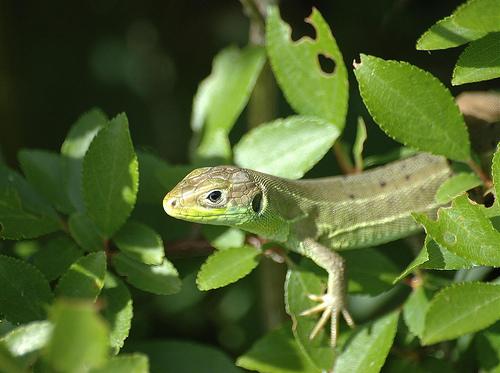}}}
        & ResNet & 
        \resizebox{0.06\textwidth}{!}{\includegraphics{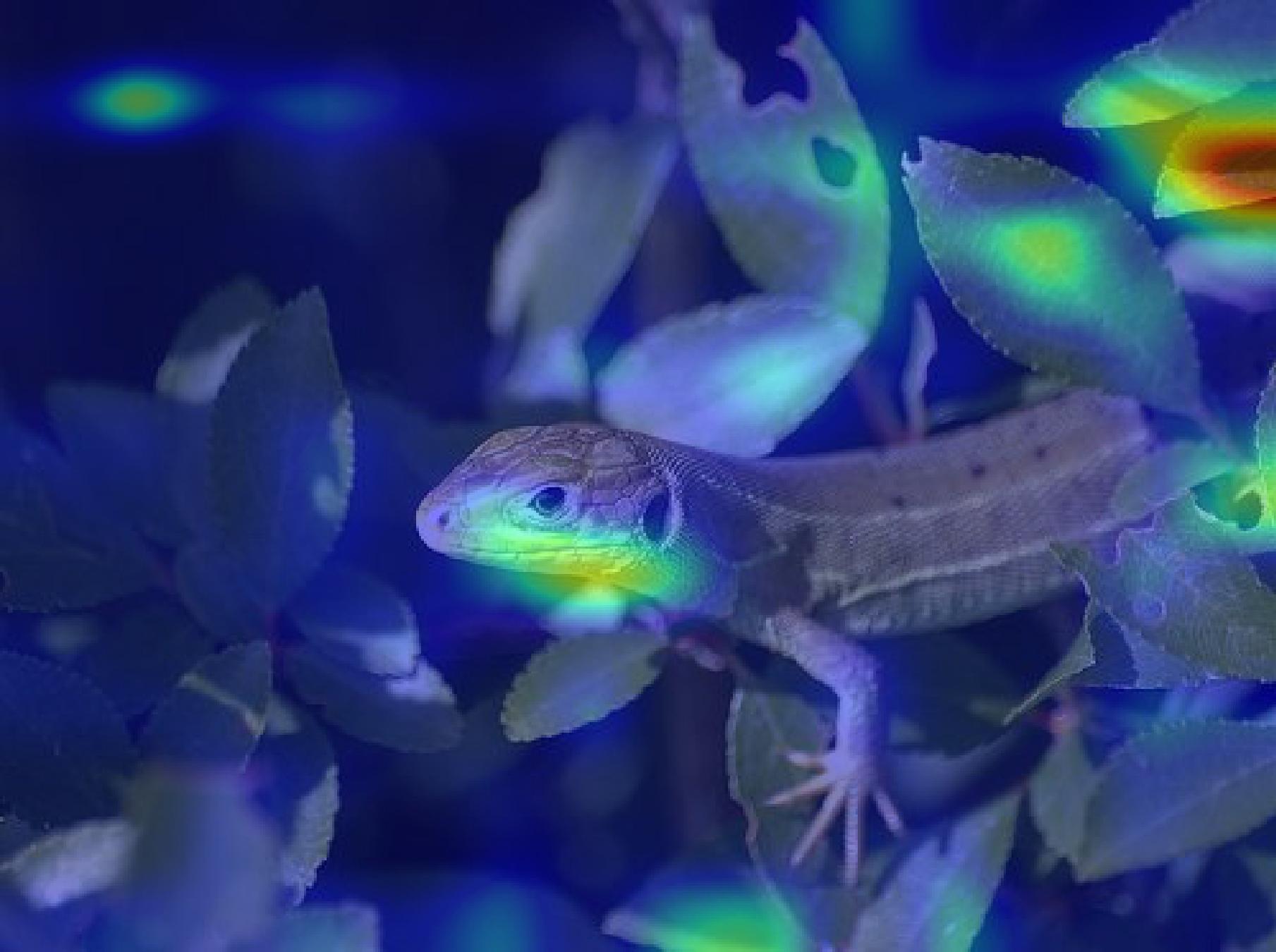}}\hspace{0mm}
        \resizebox{0.06\textwidth}{!}{\includegraphics{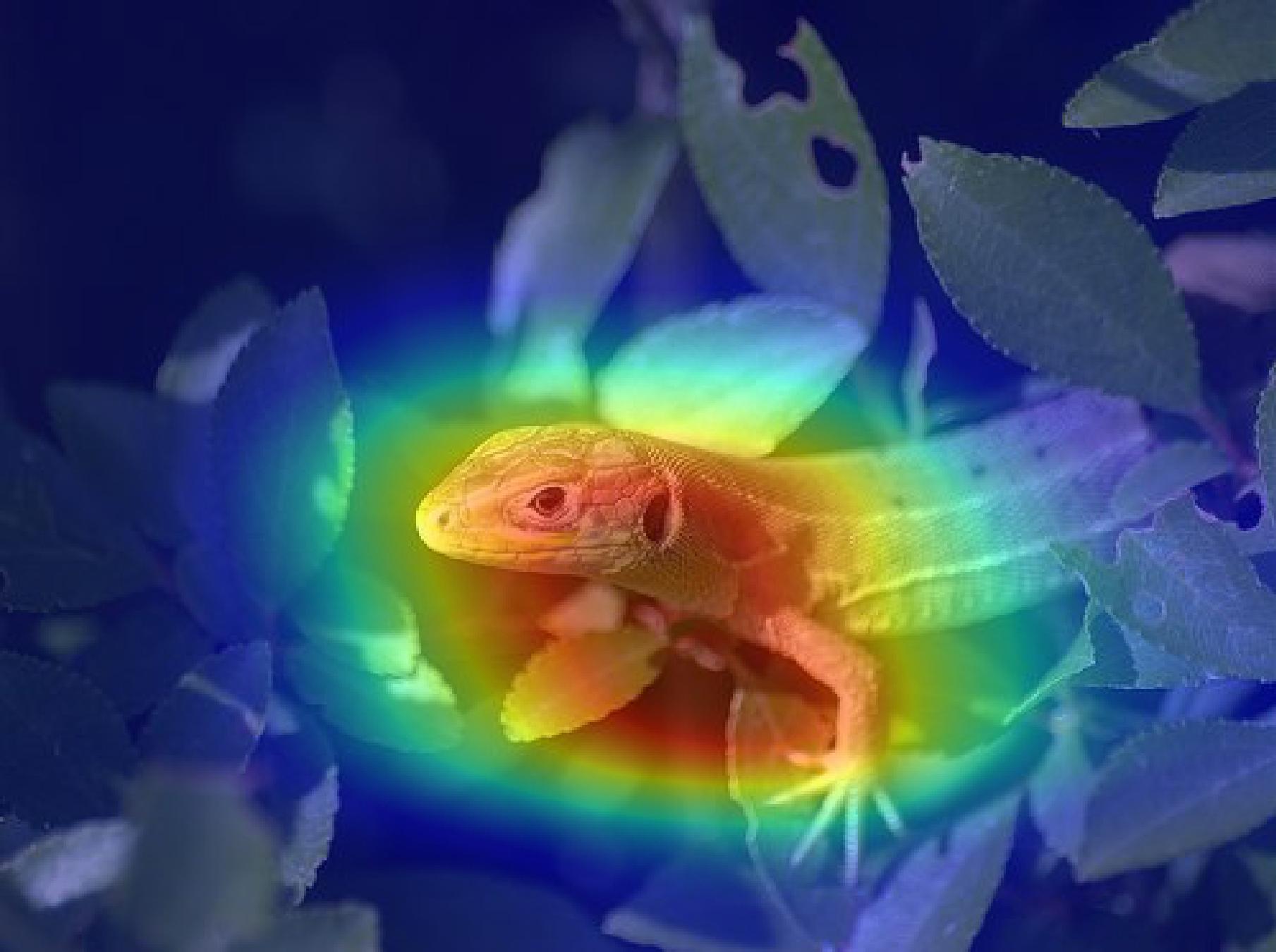}} & 0.50 \\
        \cline{2-4}
        & DAS &
        \resizebox{0.06\textwidth}{!}{\includegraphics{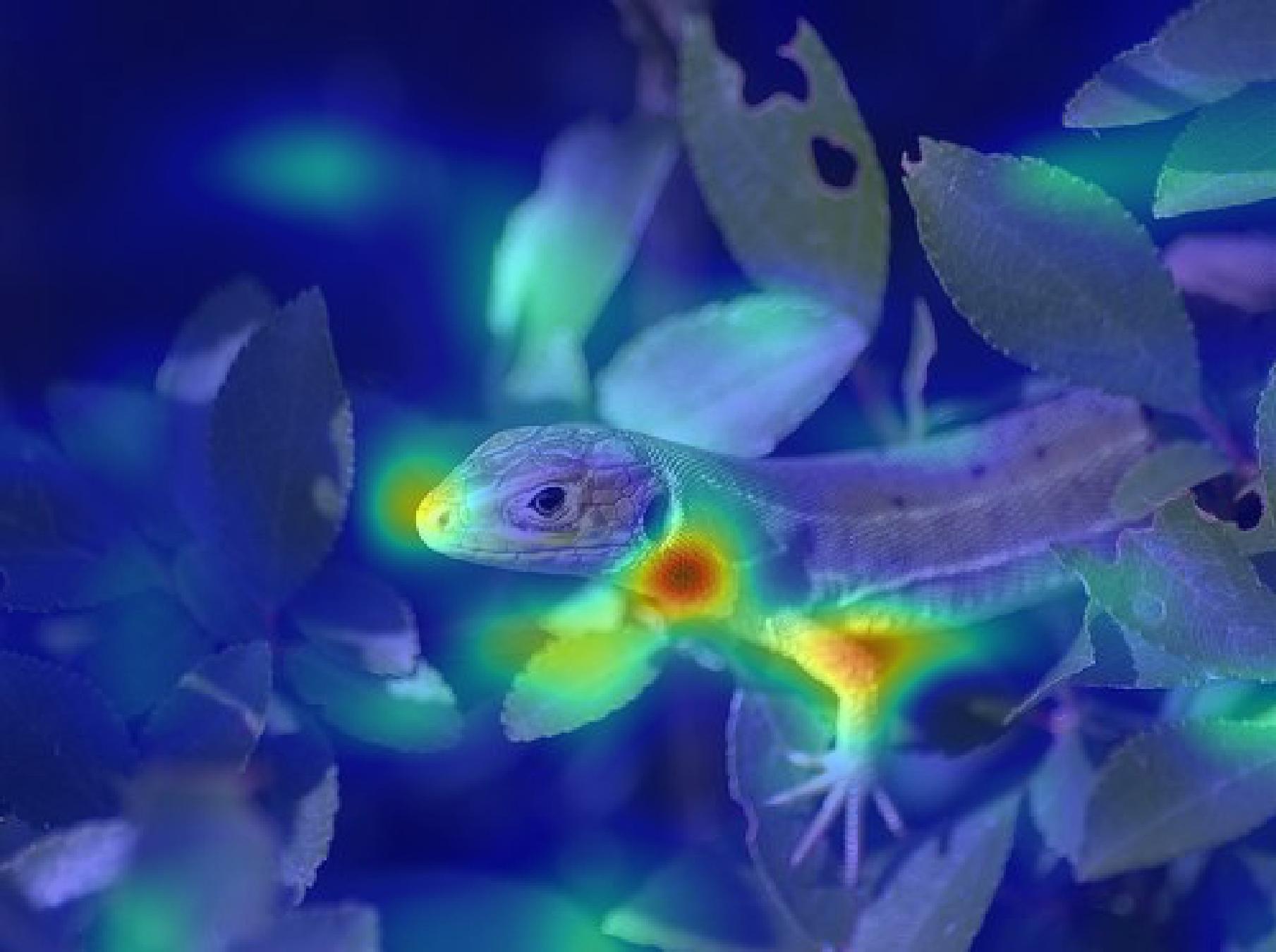}}\hspace{0mm}
        \resizebox{0.06\textwidth}{!}{\includegraphics{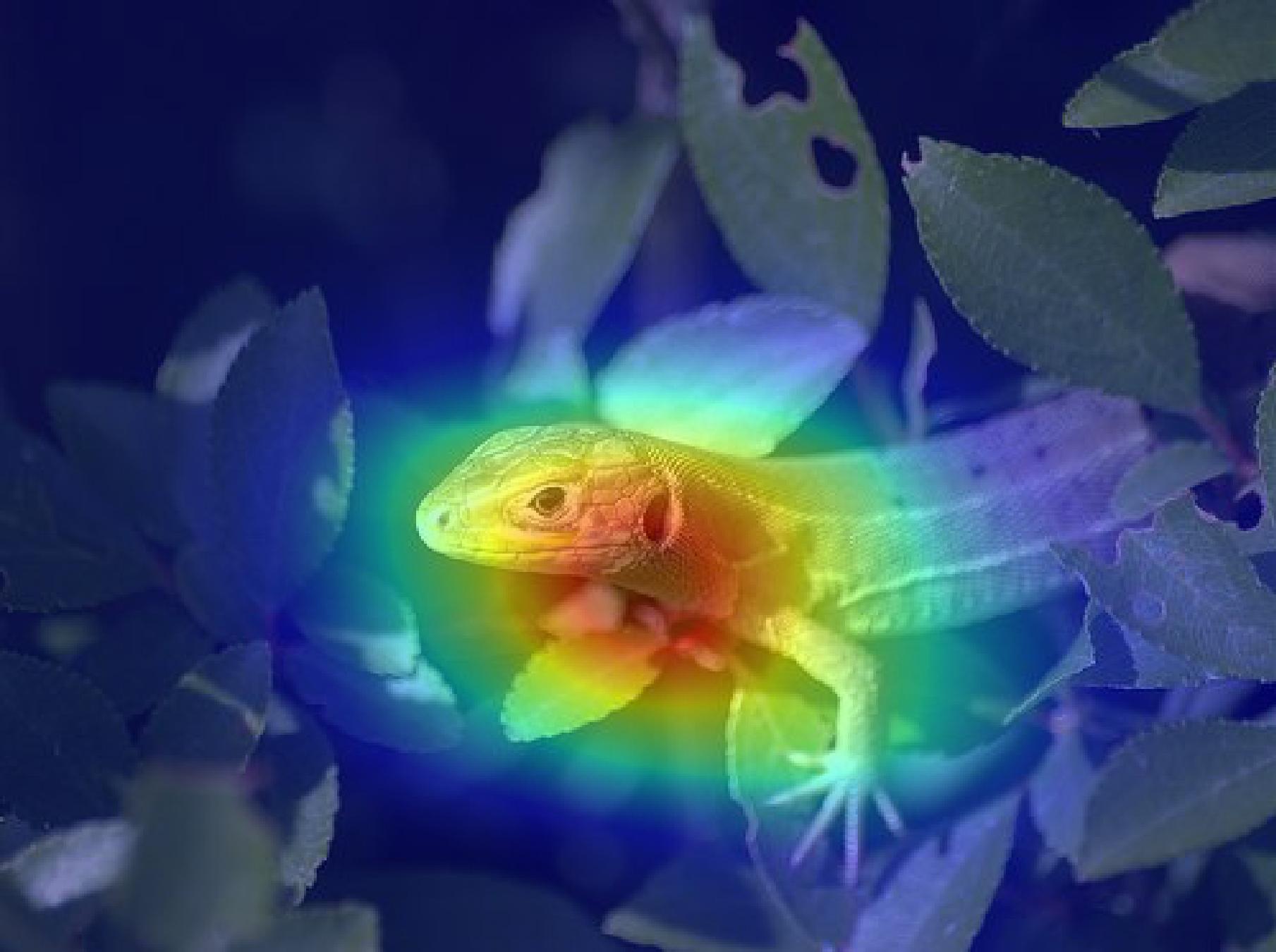}} & \textbf{0.64} \\

        \hline
        
        \multirow{3}{*}{\resizebox{0.10\textwidth}{!}{\includegraphics{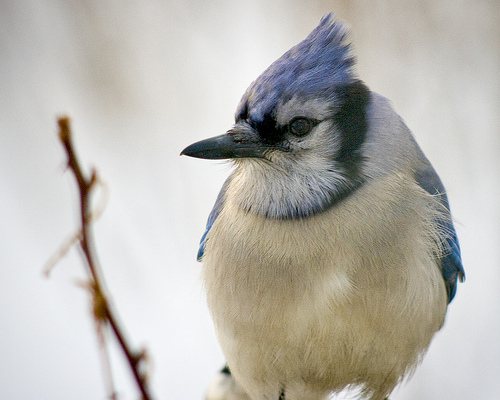}}}
        & ResNet & 
        \resizebox{0.06\textwidth}{!}{\includegraphics{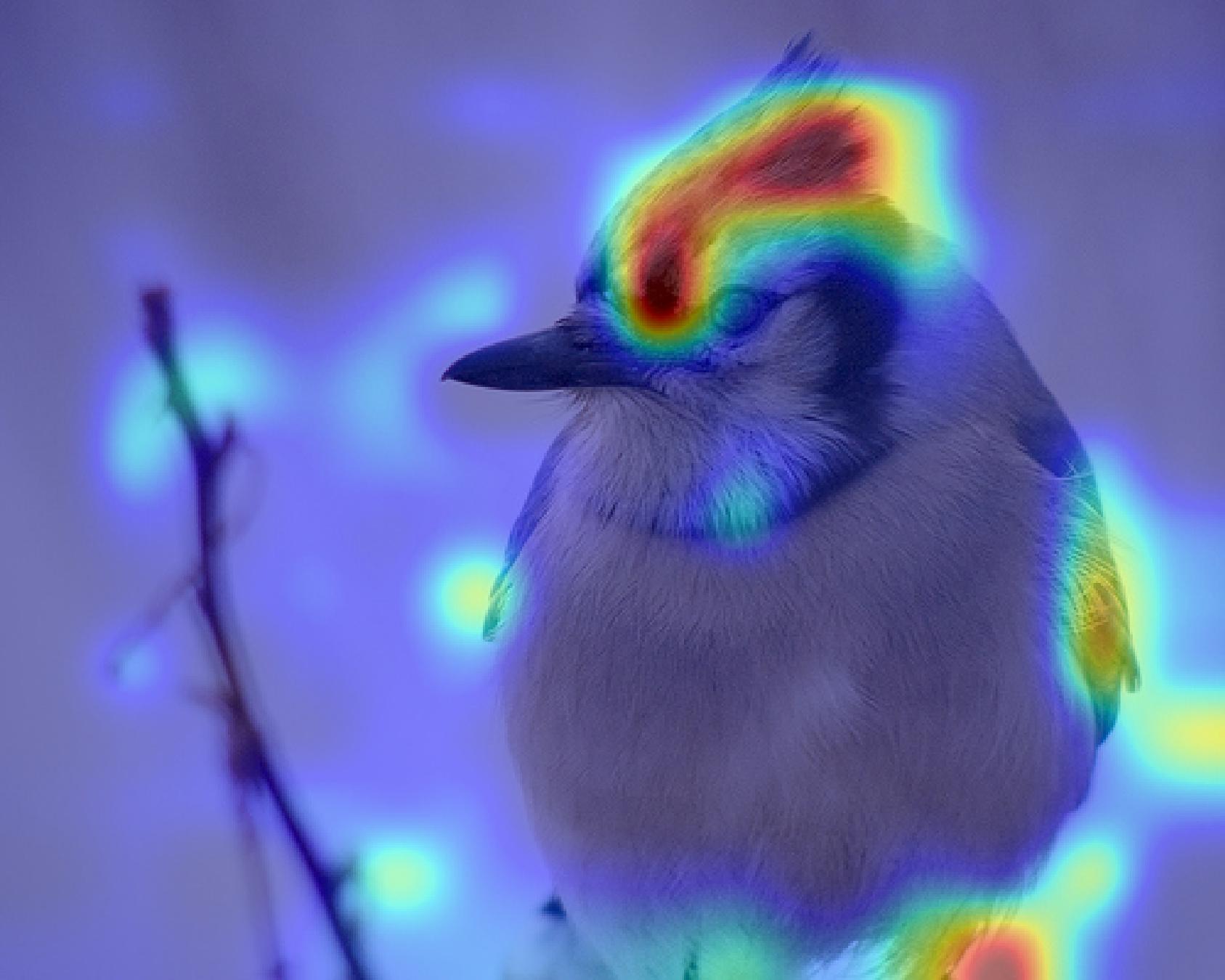}}\hspace{0mm}
        \resizebox{0.06\textwidth}{!}{\includegraphics{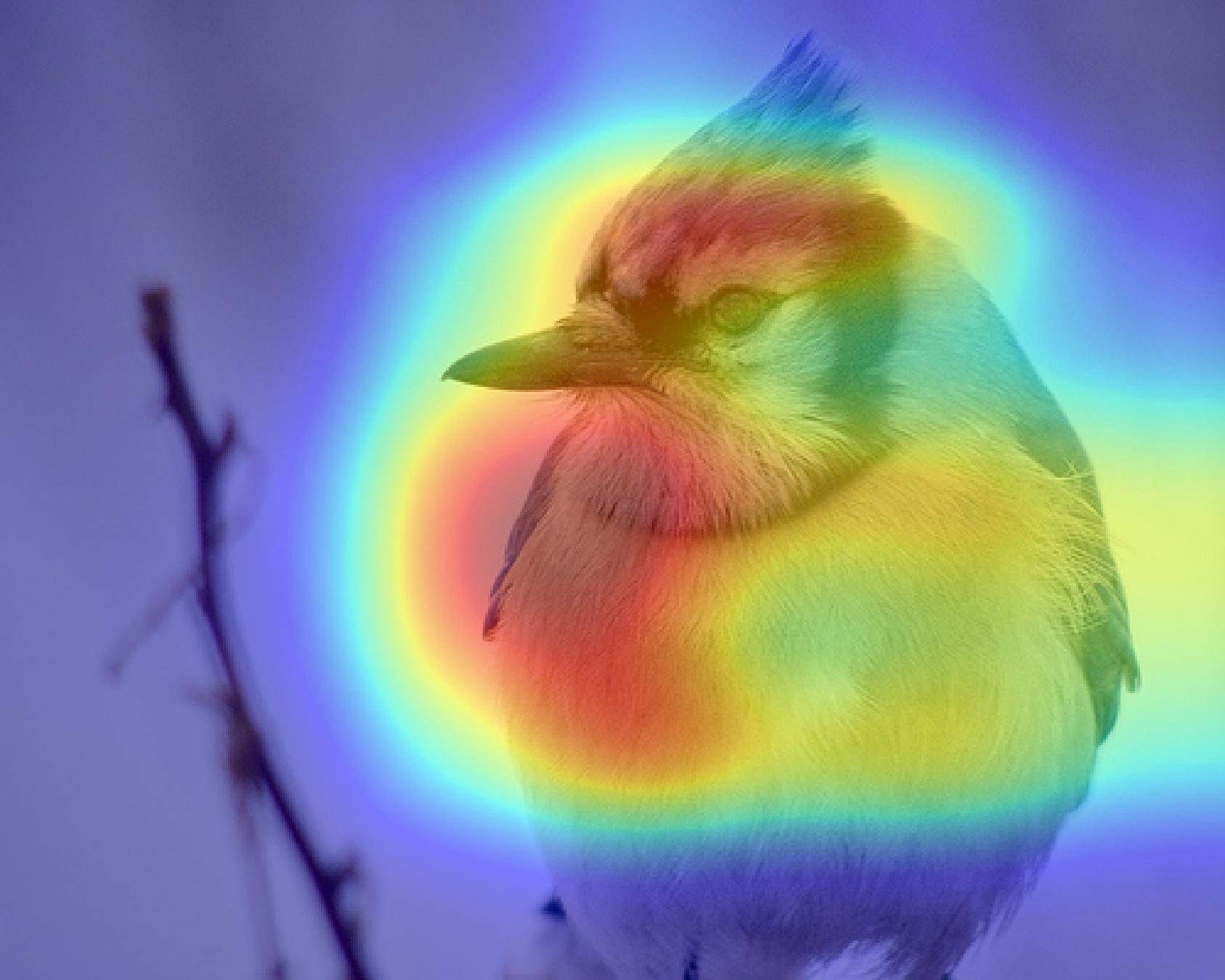}} & 0.53 \\
        \cline{2-4}
        & DAS &
        \resizebox{0.06\textwidth}{!}{\includegraphics{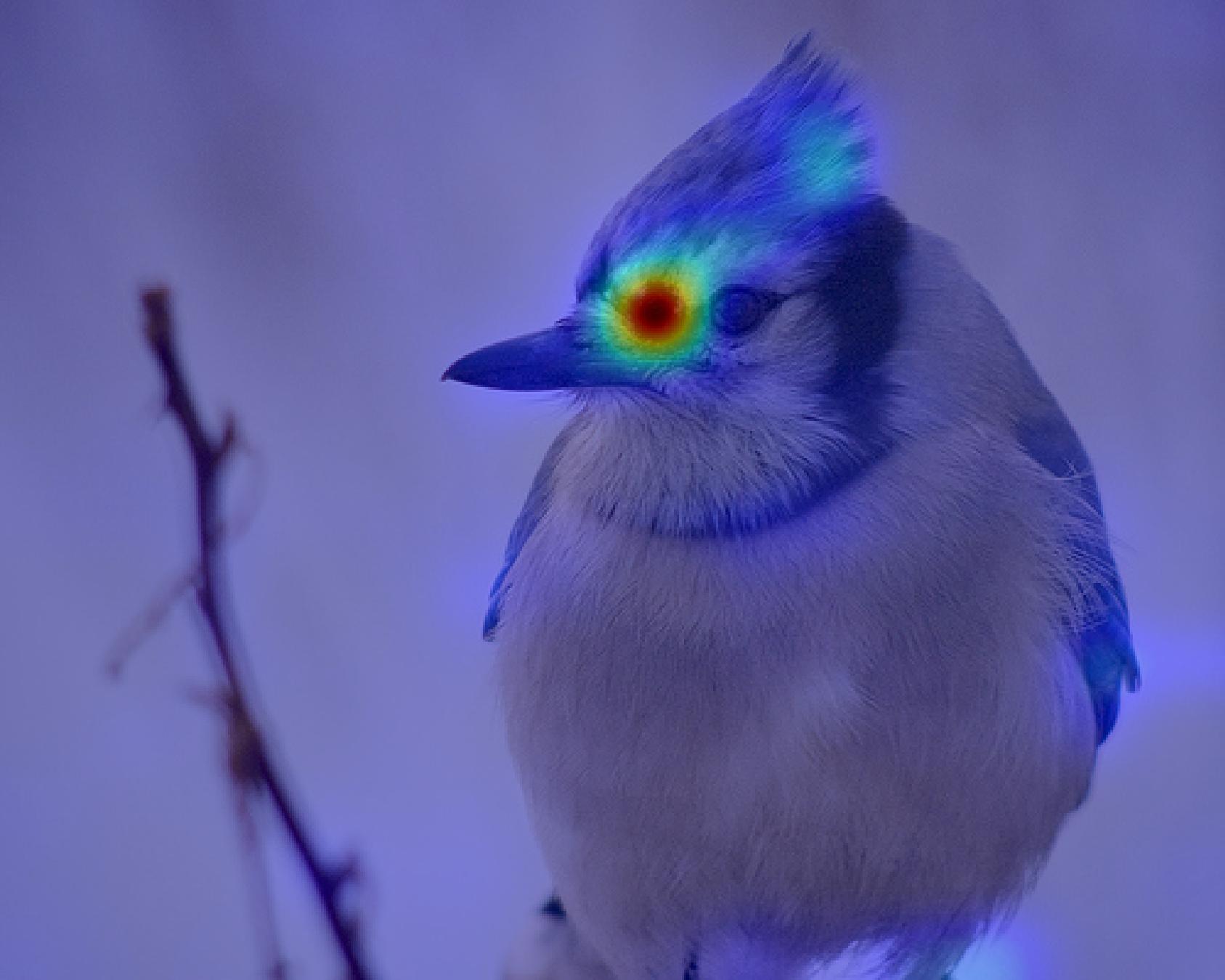}}\hspace{0mm}
        \resizebox{0.06\textwidth}{!}{\includegraphics{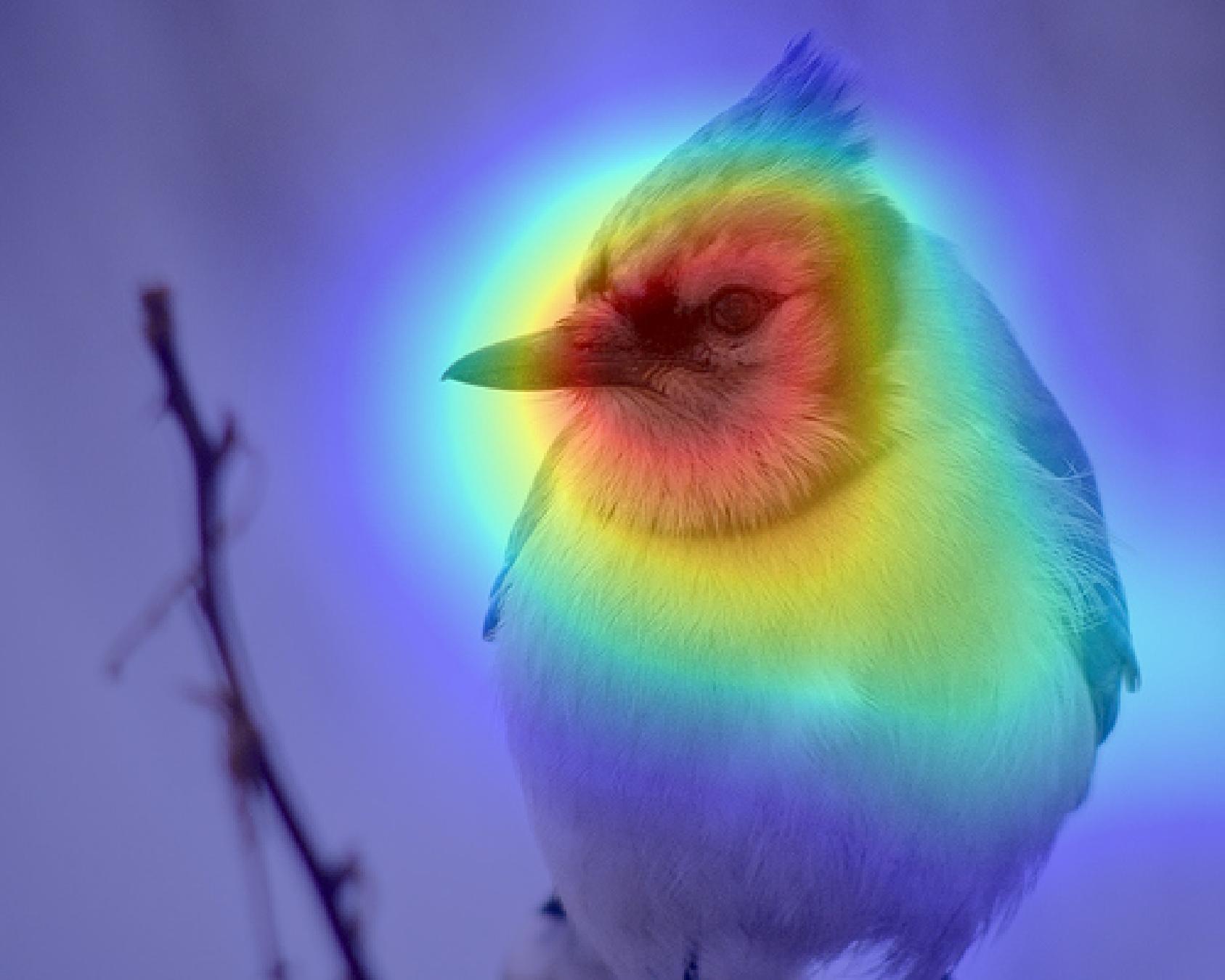}} & \textbf{0.91} \\
        
        \bottomrule
    \end{tabular}
    \caption{Analyzing GradCam activations in ResNet and DAS in Blocks 3 (left), and 4 (right), showcasing the superior saliency concentration of our method. DAS achieves a higher $sfd$ metric (\ref{sfd}), emphasizing its capability for attending to salient image features.}
    \label{fig:GradCam}
\end{figure}

Figure \ref{fig:GradCam} shows the heatmaps after block 3 and block 4 for a number of samples for ResNet-50 with and without the attention gate. These cases clearly show that our attention gate is better at focusing attention on relevant features in the image.

%% Tabulated figure here with some heatmap images

We have applied our attention gate at the end of each block in ResNet, so that the network starts focusing attention on relevant features in the early stages as well. Observing the change in heatmaps in Fig. \ref{fig:GradCam} from block 3 to block 4, we can see that attention does indeed shift towards relevant features when using DAS attention.
%% Tabulated figure here with some heatmap images

%Baldock et al. [xx] introduced the concept of prediction depth of a sample which refers to the earliest layer, after which the layer-wise k-nearest neighbor (k-NN) probe of all layers is the same as the network's prediction. We computed prediction depth for ... and our results show that our attention gate improves prediction depth.
%% Tabulated values fpr prediction depth here. 

Lastly, we define a simple metric for the effectiveness of a trained network in focusing on relevant features. We base it on weights output by gradCAM. Since we observed that gradCAM weights are compressed within the range 0 to 1, we use antilog scaling of gradCAM weights in the following. 
Let $R$ denotes the region(s) containing task-relevant features ideally identified by a human, but could also be approximated using a visual grounding tool. 
$B$ denotes the bounding box within the image which contains $R$, and is such that weights outside $B$ are low (below a threshold), that is features deemed unimportant by the network are outside B. $W_r$ denotes the average weight of features in $R$.
$W_n$ is the average weight of features in $B-R$. 
Salient feature detection score is,
\begin{equation} \label{sfd}
    sfd =  W_r/(W_r + W_n)
\end{equation}
$W_r/W_n$ provides a measure of the strength of attention paid to relevant features in the image. The higher its value, the more attention is paid to relevant features. On the other hand, a high value for $W_n/W_r$ implies that attention is being given to irrelevant features. $sfd$ will vary from 0 to 1. A score closer to 1 implies focused attention to relevant features and a score closer to 0 implies completely misplaced attention. In-between values indicate that attention is spread over relevant and irrelevant features. We use the following procedure for detecting R and B. We first use Grounding-DINO+SAM \cite{liu2023grounding, kirillov2023segment} to identify the object to be classified in an image. To avoid manual checking, we accept the possible error in this operation. This gives us the region R of relevant features. Outside of R we select the region which as per gradCAM contains salient pixels. This along with R gives us B. The last column in Fig. \ref{fig:GradCam} has $sfd$ values computed for ResNet-50 and DAS. We also computed $sfd$ values for a random sample of 100 images from ImageNet. The $sfd$ for ResNet and DAS are 0.59 and 0.72, respectively, illustrating the strength of our method in achieving targeted feature attention.

%The table below shows values for xxx.

\setlength{\tabcolsep}{5pt}
